\documentclass[10pt,journal,compsoc]{IEEEtran}

\ifCLASSOPTIONcompsoc
  \usepackage[nocompress]{cite}
\else
  \usepackage{cite}
\fi
\usepackage{diagbox}
\usepackage{graphicx}
\usepackage{ragged2e}
\usepackage{multirow}
\usepackage{amssymb}
\usepackage{amsmath}
\usepackage{booktabs}
\usepackage{algorithmic}
\usepackage[switch]{lineno}
\usepackage[dvipsnames,svgnames,table]{xcolor}
\usepackage{amsmath,amssymb}
\usepackage[ruled,linesnumbered]{algorithm2e}

\usepackage{bm}
\usepackage{multirow}
\usepackage{bbding}
\usepackage{color}
\usepackage{pifont}  % ✔ & ×
\SetArgSty{textnormal}
\SetKwInOut{Input}{Input}
\SetKwInOut{Output}{Output}
\DontPrintSemicolon

\makeatletter
\newcommand{\setalgotoprulecolor}[1]{\colorlet{toprulecolor}{#1}}
\let\old@algocf@pre@ruled\@algocf@pre@ruled % Adjust top rule colour
\renewcommand{\@algocf@pre@ruled}{\textcolor{toprulecolor}{\old@algocf@pre@ruled}}

\newcommand{\setalgobotrulecolor}[1]{\colorlet{bottomrulecolor}{#1}}
\let\old@algocf@post@ruled\@algocf@post@ruled % Adjust middle rule colour
\renewcommand{\@algocf@post@ruled}{\textcolor{bottomrulecolor}{\old@algocf@post@ruled}}

\newcommand{\setalgomidrulecolor}[1]{\colorlet{midrulecolor}{#1}}
\renewcommand{\algocf@caption@ruled}{%
  \box\algocf@capbox{\color{midrulecolor}\kern\interspacetitleruled\hrule
    width\algocf@ruledwidth height\algotitleheightrule depth0pt\kern\interspacealgoruled}}
\makeatother

\setalgotoprulecolor{blue}% Default
\setalgobotrulecolor{blue}% Default
\setalgomidrulecolor{blue}% Default
\usepackage{enumitem}
\usepackage{color}
\usepackage{ragged2e}
\usepackage{hyperref}
\usepackage{cite}
\usepackage{array}
\usepackage{colortbl}
\usepackage{subcaption}
\usepackage[export]{adjustbox}
\usepackage{wrapfig}

\ifCLASSINFOpdf
\else
\fi

\renewcommand{\justify}{\leftskip=0pt \rightskip=0pt plus 0cm}

\begin{document}

\title{AHCQ-SAM: Toward Accurate and Hardware-Compatible Post-Training Segment Anything Model Quantization}

\author{Wenlun Zhang, Yunshan Zhong, Weiqi Yan, Shengchuan Zhang, Shimpei Ando, Kentaro Yoshioka
\IEEEcompsocitemizethanks{

\IEEEcompsocthanksitem W. Zhang, S. Ando, and K. Yoshioka are with the Department of Electronics and Electrical Engineering, Keio University, Kanagawa 223-8522, Japan.
\IEEEcompsocthanksitem Y. Zhong (Corresponding  Author) is with the School of Computer Science and Technology, Hainan University, Hainan 570228, China (e-mail: yszhong01@gmail.com).
\IEEEcompsocthanksitem W. Yan and S. Zhang are with the Key Laboratory of Multimedia Trusted Perception and Efficient Computing, Ministry of Education of China, Xiamen University, Xiamen 361005, China.%.
}
}

% The paper headers
\markboth{IEEE TRANSACTIONS ON PATTERN ANALYSIS AND MACHINE INTELLIGENCE UNDER REVIEW}%
{Shell \MakeLowercase{\textit{et al.}}: Bare Demo of IEEEtran.cls for IEEE Journals}

\IEEEtitleabstractindextext{%
\begin{abstract}
\justify{
The Segment Anything Model (SAM) has revolutionized image and video segmentation with its powerful zero-shot capabilities. However, its massive parameter scale and high computational demands hinder efficient deployment on resource-constrained edge devices. While Post-Training Quantization (PTQ) offers a practical solution, existing methods still fail to handle four critical quantization challenges: (1) ill-conditioned weights; (2) skewed and long-tailed post-GELU activations; (3) pronounced inter-channel variance in linear projections; and (4) exponentially scaled and heterogeneous attention scores. To mitigate these bottlenecks, we propose AHCQ-SAM, an accurate and hardware-compatible PTQ framework featuring four synergistic components: (1) Activation-aware Condition Number Reduction (ACNR), which regularizes weight matrices via a proximal point algorithm to suppress ill-conditioning; (2) Hybrid Log-Uniform Quantization (HLUQ), which combines power-of-two and uniform quantizers to capture skewed post-GELU activations; (3) Channel-Aware Grouping (CAG), which clusters channels with homogeneous statistics to achieve high accuracy with minimal hardware overhead; and (4) Logarithmic Nonlinear Quantization (LNQ), which utilizes logarithmic transformations to adaptively adjust quantization resolution for exponential and heterogeneous attention scores. Experimental results demonstrate that AHCQ-SAM outperforms current methods on SAM. Compared with the SOTA method, it achieves a 15.2\% improvement in mAP for 4-bit SAM-B with Faster R-CNN on the COCO dataset. Furthermore, we establish a PTQ benchmark for SAM2, where AHCQ-SAM yields a 14.01\% improvement in $\mathcal{J}\&\mathcal{F}$ for 4-bit SAM2-Tiny on the SA-V Test dataset. Finally, FPGA-based implementation validates the practical utility of AHCQ-SAM, delivering a 7.12$\times$ speedup and a 6.62$\times$ power efficiency improvement over the floating-point baseline. \textbf{Code is available at} \href{https://github.com/Wenlun-Zhang/AHCQ-SAM}{\textit{https://github.com/Wenlun-Zhang/AHCQ-SAM}.}

}
\end{abstract}

% Note that keywords are not normally used for peerreview papers.
}

% make the title area
\maketitle

%\IEEEdisplaynontitleabstractindextext

\IEEEpeerreviewmaketitle

%%%%%%%%% BODY TEXT
\section{Introduction}
\label{sec:intro}
\IEEEPARstart{T}{h}e Segment Anything Model (SAM)~\cite{Segment_Anything,ravisam} is a powerful tool for image/video segmentation, demonstrating strong zero-shot performance across diverse visual domains and broad applicability in real-world scenarios~\cite{Sam-med2d,Personalize_SAM,SAM_Shadow_Detction,Follow_Anything,Anything-3d}. However, its large-scale parameters, substantial storage demands, and high computational costs pose significant challenges for deployment on edge devices~\cite{ViT_Compression_Gaussian,Less_is_more,I-vit,Post-training_ViT}.

\begin{figure*}[htbp]
\centering
\begin{subfigure}{0.45\linewidth}
    \centering
    \includegraphics[width=\linewidth]{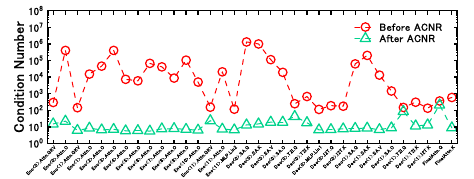}
    \caption{Challenge 1: Numerous weight matrices exhibit severe numerical ill-conditioning.}
    \label{sam_cond_comparison}
\end{subfigure}
\begin{subfigure}{0.45\linewidth}
    \centering
    \includegraphics[width=\linewidth]{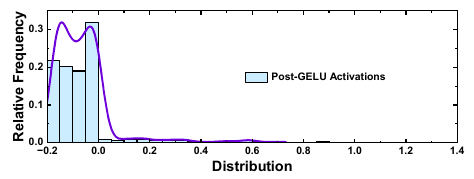}
    \caption{Challenge 2: Highly-skewed and long-tailed post-GELU activations (the 10th block of the image encoder).}
    \label{Fig_Post_GELU_Acts}
\end{subfigure}

\begin{subfigure}{0.45\linewidth}
        \includegraphics[width=\linewidth]{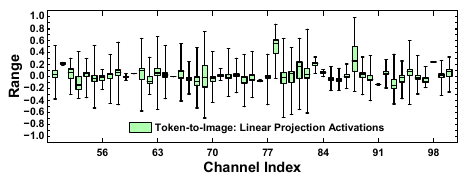}
        \caption{Challenge 3: Highly variable inter-channel activations (channel 50 to 100 of the first block of the mask decoder).} 
        \label{Fig_Lin_Proj_Acts}
\end{subfigure}
\begin{subfigure}{0.45\linewidth}
        \includegraphics[width=\linewidth]{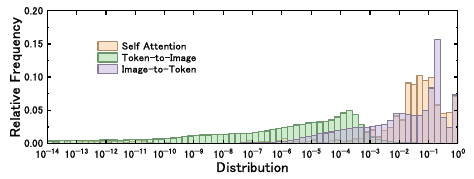}
        \caption{Challenge 4: Exponentially scaled and heterogeneous post-Softmax activations.} 
        \label{Fig_Softmax_Dist}
\end{subfigure}
\caption{Challenges in SAM quantization (Data obtained from the SAM-B model with the YOLOX detector).}
\label{Fig_Acts_Challenges}
\end{figure*}

To tackle this challenge, model quantization has been widely adopted to replace floating-point weights and activations with low-bit representations, reducing storage overhead and enabling efficient integer-based computations, making it well-suited for edge devices with constrained resources~\cite{Quant_Whitepaper}. One prominent approach is Quantization-Aware Training (QAT), which incorporates quantization effects during training, allowing the model to adapt to quantized weights and activations~\cite{LSQ,Q-vit,Power-of-Two_Quant,Differentiable_Soft_Quant,Oscillation-free_Quant}. However, applying QAT to SAM is impractical due to the computational expense of training. For instance, the training of SAM relies on a co-developed data engine and consequently utilizes the SA-1B dataset comprising 1.1B masks and 11M images~\cite{Segment_Anything}. As a more practical alternative, Post-Training Quantization (PTQ) has gained increasing attention. PTQ requires only a small calibration dataset, significantly reducing data and computational demands while maintaining competitive accuracy~\cite{Similarity_Aware_Quant,Fq-vit,Noisyquant,Quant_loss_landscape,Trio-ViT,ERQ,ERQ_PAMI,zhong2023s,APQ-ViT}. 

Recent works have demonstrated the feasibility of applying PTQ to SAM~\cite{DBLP:conf/ijcai/RenLW0Y25,zhang2025saq,PTQ4SAM,PQ-SAM,Ranjan_2025_CVPR,Tinysam}. For example, PTQ4SAM~\cite{PTQ4SAM} utilizes equivalent sign transformations and adaptive resolution quantization to accommodate SAM’s unique activation distributions. PQ-SAM~\cite{PQ-SAM} incorporates a grouped activation distribution transformation that hierarchically clusters and adjusts activation channels. Despite these advancements, our study reveals that existing PTQ methods suffer from intolerable performance degradation, limiting their practical utility for ultra-low-bit deployment and motivating the need for more effective quantization techniques.

In this paper, we identify four critical challenges that limit the efficacy of PTQ for SAM. First, numerous weight matrices in SAM layers exhibit severe numerical ill-conditioning (red curve of Fig.~\ref{sam_cond_comparison}). These excessively high condition numbers significantly exacerbate the model's sensitivity to quantization-induced perturbations~\cite{liu2025condiquant}. Second, post-GELU activations (Fig.~\ref{Fig_Post_GELU_Acts}) manifest highly skewed and long-tailed distributions, which present a substantial mismatch for conventional hardware-friendly uniform or power-of-two quantizers. Third, activations in \texttt{Query}/\texttt{Key}/\texttt{Value} projections and the first \texttt{Linear} layer of MLP exhibit pronounced inter-channel variance (Fig.~\ref{Fig_Lin_Proj_Acts}), necessitating a more sophisticated quantization granularity beyond standard per-tensor quantization. Fourth, post-Softmax attention scores (Fig.~\ref{Fig_Softmax_Dist}) display an exponentially scaled range and highly heterogeneous distribution patterns, requiring a distribution-flexible quantizer capable of adapting to rapidly shifting numerical ranges.

\begin{figure}[htbp]
    \centering
    \includegraphics[width=\linewidth]{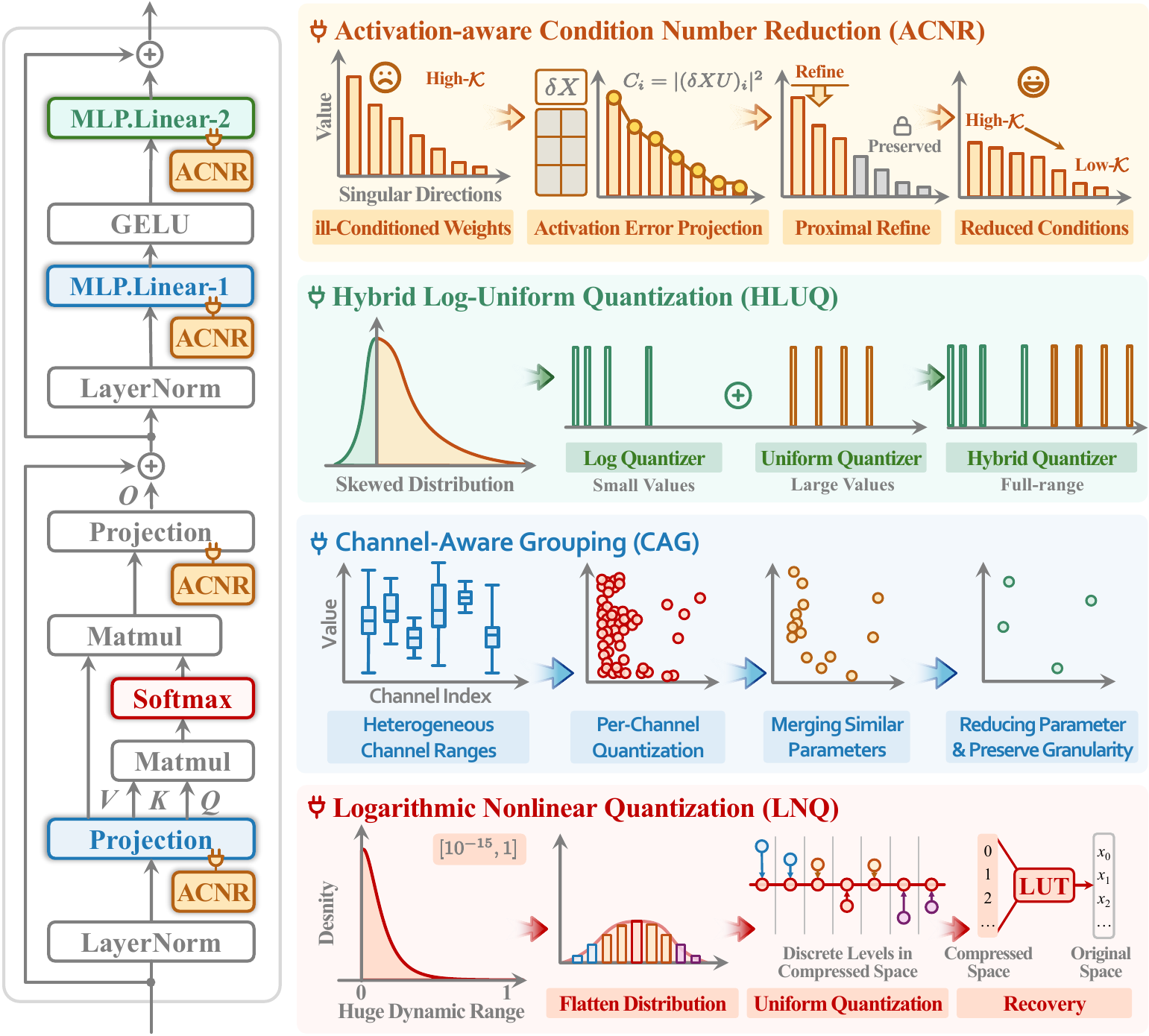}
    \caption{AHCQ-SAM framework: ACNR regularizes the ill-conditioned weight matrices, HLUQ refines quantization resolution for skewed and long-tailed post-GELU activations, CAG groups parameters to manage highly variable inter-channel activations, and LNQ utilizes logarithmic transformation to accommodate exponentially scaled and heterogeneous post-Softmax activations.}
    \label{Fig_Framework}
\end{figure}

To mitigate these identified challenges, we propose AHCQ-SAM, an accurate and hardware-compatible PTQ framework for SAM. As shown in Fig.~\ref{Fig_Framework}, AHCQ-SAM synergistically integrates four novel components, including Activation-aware Condition Number Reduction (ACNR), Hybrid Log-Uniform Quantization (HLUQ), Channel-Aware Grouping (CAG), and Logarithmic Nonlinear Quantization (LNQ), each of which targets a specific quantization challenge. Specifically, to stabilize ill-conditioned weights, ACNR leverages a proximal point algorithm to enhance activation-aware stability. By incorporating the empirical distribution of activation quantization errors, it selectively regularizes the weight matrices' condition numbers, specifically suppressing the feature directions that are vulnerable to quantization errors. To handle the unique post-GELU activations, HLUQ innovatively applies a power-of-two quantizer for densely clustered small values and a uniform quantizer for sparse but widely-distributed large values, effectively capturing the heavy-tailed nature of post-GELU activations while maintaining hardware efficiency. Furthermore, to tackle pronounced inter-channel variance, CAG selectively clusters channels with homogeneous statistical properties, achieving accuracy comparable to per-channel quantization while supporting hardware-friendly implementation by reducing on-chip register overhead by 99.7\%. Finally, to accommodate the exponentially scaled and heterogeneous attention scores, LNQ utilizes a logarithmic transformation to adaptively adjust the quantization resolution for infinitesimal scores while compressing high-magnitude values. It further leverages redundant on-chip BRAM resources as Look-Up Tables (LUTs), enabling efficient value transformation while avoiding complex arithmetic computations.

By integrating these innovations, AHCQ-SAM significantly reduces accuracy degradation at 5-bit precision and improves 4-bit performance by a large margin. Furthermore, we extend AHCQ-SAM to establish a PTQ benchmark for SAM2~\cite{ravisam}, which targets Video Object Segmentation (VOS) tasks. Experimental results demonstrate that AHCQ-SAM consistently outperforms existing state-of-the-art methods across all metrics. For instance, AHCQ-SAM improves the mAP by 15.2\% over PTQ4SAM for 4-bit SAM-B with Faster R-CNN on the COCO dataset, and increases the $\mathcal{J}\&\mathcal{F}$ score by 14.01\% for 4-bit SAM2-Tiny on the SA-V Test dataset. Finally, we validate the practical effectiveness of AHCQ-SAM by implementing it on an FPGA-based accelerator, demonstrating significant gains in both processing speed and power efficiency. Our primary contributions are summarized as follows:

\begin{itemize}[leftmargin=*]
    \item We systematically identify four critical challenges in SAM quantization: (1) ill-conditioned weights causing sensitivity to quantization errors; (2) skewed and long-tailed post-GELU activations; (3) pronounced inter-channel variance in linear projections; and (4) exponentially-ranged and heterogeneous attention scores.

    \item To address these challenges, we propose AHCQ-SAM, featuring four synergistic components, including ACNR, HLUQ, CAG, and LNQ, each specifically designed to mitigate one identified challenge while maintaining hardware compatibility.

    \item Extensive experiments show that AHCQ-SAM consistently achieves superior performance. For instance, it yields a 15.2\% mAP improvement over PTQ4SAM for the 4-bit SAM-B with Faster R-CNN on the COCO dataset. Furthermore, we establish a PTQ benchmark for SAM2, where AHCQ-SAM sets a strong baseline by improving the $\mathcal{J}\&\mathcal{F}$ score by 14.01\% for 4-bit SAM2-Tiny on the SA-V Test dataset.
    
    \item We further develop an FPGA-based accelerator to evaluate AHCQ-SAM’s hardware efficiency. Our results indicate that AHCQ-SAM delivers a $7.12\times$ speedup and $6.62\times$ power efficiency improvement over floating-point implementations, demonstrating superior resource utilization.
    
\end{itemize}

\section{Related Work}

\subsection{Efficient SAM}

The substantial computational overhead of SAM remains a significant bottleneck for deployment on edge devices. Consequently, a variety of efficient SAM methods have emerged to strike a balance between segmentation accuracy and resource efficiency~\cite{DBLP:journals/ijcv/SunLSZH25}. To this end, model compression techniques like knowledge distillation~\cite{zhou2025edgesam,zhang2024efficientvit,mobile_sam,DBLP:conf/aaai/ShuLTZCL0025,ke2023segment}, model pruning~\cite{chen2024slimsam,abebe2025supersam}, and feature caching~\cite{zhang2026efficient} have been extensively explored. Among distillation-based methods, MobileSAM~\cite{mobile_sam} adopts decoupled distillation to replace the heavy image encoder with a lightweight counterpart, while TinySAM~\cite{DBLP:conf/aaai/ShuLTZCL0025} leverages a full-stage distillation pipeline coupled with hard prompt sampling and a hierarchical segmenting strategy. Regarding pruning, SlimSAM~\cite{chen2024slimsam} introduces an alternate slimming framework that progressively prunes and distills decoupled sub-structures, enhancing knowledge inheritance under extreme pruning ratios and limited data. Furthermore, feature caching approaches like Efficient-SAM2~\cite{zhang2026efficient} optimize computation via object-aware sparse window routing and memory retrieval, effectively filtering background redundancy. However, these methods still rely on expensive full-precision computation.

\subsection{Post-training Quantization}

Post-training Quantization (PTQ) utilizes a small calibration dataset to determine quantization parameters. Compared with quantization-aware training (QAT), it enables rapid deployment on edge devices without requiring extensive retraining. AdaRound~\cite{AdaRound} identifies the sensitivity of weight rounding and introduces an optimization technique to reduce overall model loss. BRECQ~\cite{Brecq} employs block reconstruction to strike a balance between cross-layer dependency and generalization error. QDrop~\cite{Qdrop} integrates dropout into the reconstruction process to improve the flatness of the optimized models. Despite their success, these methods are primarily designed for CNN-based models and face challenges when applied to Transformer architectures. 

In the domain of Transformer-based models, FQ-ViT~\cite{Fq-vit} improves granularity using powers-of-two scale and Log-Int-Softmax while maintaining hardware efficiency. RepQ-ViT~\cite{Repq-vit} eliminates parameter overhead in per-channel quantization by applying reparameterization techniques to post-LayerNorm activations. Evol-Q~\cite{frumkin2023jumping} adopts an evolutionary search to determine the disturbance-sensitive quantization parameters. To smooth the optimization, Bit-shrinking~\cite{lin2023bit} introduces a sharpness term and a self-adapted bit-shrinking scheduler that gradually reduces bit-widths. ERQ~\cite{ERQ_PAMI,ERQ} minimizes quantization errors via ridge regression, while PTQ4ViT~\cite{Ptq4vit} employs a twin uniform quantizer to effectively handle post-Softmax and post-GELU activations. DopQ-ViT~\cite{yang2024dopq} introduces a Tan quantizer to better preserve the power-law distribution of post-Softmax activations and a MAD-guided optimal scaling factor to mitigate the influence of outliers in post-LayerNorm layers. IGQ-ViT~\cite{moon2024instance} employs instance-aware group quantization, where activations are split into multiple groups dynamically for each instance. In OAS-ViT~\cite{maoutlier}, theoretical insights are presented to analyze reconstruction granularity and outliers within models. To tackle the accuracy degradation in ultra-low bit, APHQ-ViT~\cite{wu2025aphq} introduces an average perturbation hessian loss for accurate importance estimation and an MLP-reconstruction scheme to handle post-GELU quantization. FIMA-Q~\cite{wu2025fima} introduces a quantization loss based on the Fisher information matrix. While these approaches offer insights for mitigating challenges in SAM, they are not easily transferable to the SAM architecture due to its unique structural complexities.

PTQ4SAM~\cite{PTQ4SAM}, the first PTQ method specialized for SAM, introduces bimodal integration and adaptive log quantization to address unique distribution challenges while maintaining hardware efficiency. PQ-SAM~\cite{PQ-SAM} addresses the performance degradation caused by asymmetric activation distributions and extreme outliers through a grouped activation distribution transformation. By employing a two-stage outlier hierarchical clustering scheme, this transformation effectively scales and shifts activation channels to create a quantization-friendly distribution. SAQ-SAM~\cite{zhang2025saq} enhances PTQ for SAM by introducing a perceptual-consistency clipping to suppress extreme attention outliers and prompt-aware reconstruction to align image features with prompt intentions via cross-attention interactions.

\section{Method}

\subsection{Preliminaries}

\subsubsection{Quantizer}

Quantization discretizes continuous values into low-bit representations, enabling efficient computation and reducing memory usage. Two widely adopted quantizers are the uniform quantizer and the power-of-two quantizer. The \textbf{uniform quantizer} divides the input range into equally spaced intervals, mapping each value to the closest quantized grid:

\begin{equation}
x_{q} = \text{clamp} \left( \left\lfloor \frac{x}{s} \right\rceil + z, 0, 2^k - 1 \right),
\label{Eq_Uniform_Quant}
\end{equation}

\begin{equation}
x \approx \hat{x} = s \cdot (x_{q} - z).
\label{Eq_Uniform_Dequant}
\end{equation}

\noindent Here, $x$ is the original floating-point input, $x_q$ is the quantized integer representation, $k$ is the bit-width, $s$ is the scale factor, and $z$ is the zero point. The rounding function $\left\lfloor \cdot \right\rceil$ ensures proper discretization. Uniform quantization is widely adopted due to its straightforward hardware implementation, allowing integer arithmetic to replace floating-point operations, leading to higher efficiency and lower computational cost. For highly skewed data distributions, the \textbf{power-of-two quantizer} provides a more effective alternative, as it assigns quantization grids based on powers of two, offering higher precision for small values:

\begin{equation}
x_{q} = \text{clamp} \left( \left\lfloor -\log_{2}\frac{x}{s} \right\rceil, 0, 2^k - 1 \right),
\label{Eq_Log_Quant}
\end{equation}

\begin{equation}
x \approx \hat{x} = s \cdot 2^{-x_{q}}.
\label{Eq_Log_Dequant}
\end{equation}

\noindent The power-of-two quantizer is particularly beneficial for hardware, as it enables multiplications to be replaced by bit shifts, improving computational speed and power efficiency.

\subsubsection{Quantization Granularity}

Quantization operates at varying levels of granularity, introducing a trade-off between computational efficiency and quantization effectiveness. The two most common approaches are per-tensor and per-channel quantization. \textbf{Per-Tensor} quantization employs a single scale and zero-point across an entire weight or activation tensor, reducing computational complexity and memory overhead. However, it struggles with large inter-channel variations, leading to suboptimal quantization performance. \textbf{Per-Channel} quantization assigns individual quantization parameters to each output channel, effectively mitigating distribution variance across channels. However, it requires storing more quantization parameters, increasing memory usage and data transfer costs.

\subsubsection{Condition Number and Quantization Error}

The recent CondiQuant~\cite{liu2025condiquant} establishes a link between the conditioning of weights and the amplification of quantization error. In particular, given a linear layer $Y=XW$, the relative error in the output $\delta Y$ induced by the quantization of activations $\delta X = X - X_q$ is bounded by the condition number $\kappa(W)$ of the $W$:
\begin{equation}
    \frac{\|\delta Y\|_2}{\|Y\|_2} \leq  \kappa(W) \frac{\|\delta X\|_2}{\|X\|_2}.
\end{equation}

This inequality suggests that numerical ill-conditioning of weights can exacerbate the sensitivity of the output to quantization error.  

\subsubsection{Block-Wise Reconstruction}

We employ block-wise reconstruction~\cite{Qdrop}, as adopted in PTQ4SAM~\cite{PTQ4SAM}, to mitigate the quantization-induced error in weight and activation quantization by minimizing the mean squared error:

\begin{equation}
\mathcal{L} = \| \mathbf{O}_\mathcal{B} - \hat{\mathbf{O}}_\mathcal{B} \|_2^2,
\label{Eq_Block_Recon}
\end{equation}

\noindent where $\mathbf{O}_\mathcal{B}$ and $\hat{\mathbf{O}}_\mathcal{B}$ represent the floating-point and quantized outputs of the $\mathcal{B}$-th block, respectively.

\subsection{AHCQ-SAM}
\label{Subsection_AHCQ-SAM}

To advance SAM quantization, we propose AHCQ-SAM, a framework specifically designed to overcome the four key challenges observed in SAM. As shown in Fig.~\ref{Fig_Framework}, AHCQ-SAM leverages Activation-aware Condition Number Reduction (ACNR), Hybrid Log-Uniform Quantization (HLUQ), Channel-Aware Grouping (CAG), and Logarithmic Nonlinear Quantization (LNQ), each targeting a specific quantization challenge.

\subsubsection{Activation-aware Condition Number Reduction}

\textbf{Challenge 1.} The \textbf{first challenge} of quantization arises from the ill-conditioning of weights. As illustrated in the red curve of Fig.~\ref{sam_cond_comparison}, many layers within SAM-B manifest excessively high condition numbers. This numerical ill-conditioning exacerbates the sensitivity of the layers to quantization perturbations, thereby posing a significant bottleneck for low-bit quantization. CondiQuant~\cite{liu2025condiquant} employs a data-agnostic Proximal Gradient Descent (PGD) framework to reduce the condition number of weights. However, CondiQuant relies on an isotropic noise assumption for activation perturbations, which overlooks the fact that activation errors $\delta X$ are highly non-uniform across channels~\cite{Repq-vit,zhong2023s,ERQ_PAMI}. Furthermore, the incorporation of gradient descent during the optimization process makes CondiQuant susceptible to overfitting and limits its overall robustness (As discussed in Sec.~\ref{sec:Comparsion between ACNR and CondiQuant}).

\textbf{Solution.} To address the above challenge, we propose Activation-aware Condition Number Reduction (ACNR). By explicitly incorporating the empirical distribution of activation quantization error $\delta X$, ACNR achieves activation-aware stability by selectively regularizing the weight matrices' condition numbers, specifically penalizing singular directions most susceptible to quantization noise. To this end, we minimize the following objective:
\begin{equation}
\begin{aligned}
    \min_{W} \mathcal{J}(W) = \lambda \sum_{i=1}^r (\sigma_i(W) - t)^2 + \beta \| \delta X \cdot W \|_F^2,
\end{aligned}
\label{eq:act-aware-condi}
\end{equation}
where $\lambda$ and $\beta$ are a balance hyperparameters. To minimize $\mathcal{J}(W)$, we employ a proximal point algorithm. Unlike CondiQuant's PGD, which interleaves gradient steps with proximal operations, our approach eliminates gradient steps to avoid optimization drift caused by noisy gradients. Starting from $W_0 = W_{\text{orig}}$, we iteratively refine the weights by solving a sequence of proximal subproblems. At iteration $k$, with the help of an auxiliary variable $Z$, we compute the next iterate as:
\begin{equation}
    W_{k+1} = \arg\min_{Z} \frac{1}{2}\|Z - W_k\|_F^2 + \mathcal{J}(Z).
\end{equation}

Substituting the definition of $\mathcal{J}(\cdot)$ from Eq.~\ref{eq:act-aware-condi} yields the subproblem:
\begin{equation}
\begin{aligned}
        \min_{Z} \Phi(Z) = \frac{1}{2}\|Z - W_k\|_F^2 & + \lambda \sum_{i=1}^r (\sigma_i(Z) - t)^2 \\ & + \beta \|\delta X Z\|_F^2.
\label{eq:act-aware-condi-Substituting}
\end{aligned}
\end{equation}
The first term $\frac{1}{2}\|Z - W_k\|_F^2$ is the proximal term, ensuring the update stays close to the current solution. To obtain a tractable solution, we restrict $Z$ to share the same singular vectors $(U, V)$ as $W_k$. Let $W_k = U \Sigma V^T$ with $\Sigma = \text{diag}(\sigma_1, \dots, \sigma_r)$. We then parameterize the solution as $Z^* = U \Sigma^* V^T$ where $\Sigma^* = \text{diag}(\sigma_1^*, \dots, \sigma_r^*)$ contains the new singular values to be optimized.

Substituting this parameterization into $\Phi(Z)$ decouples the optimization into $r$ independent scalar subproblems:
\begin{equation}
\min_{\sigma_i^*} \mathcal{L}(\sigma_i^*) = \frac{1}{2}(\sigma_i^* - \sigma_i)^2 + \lambda(\sigma_i^* - t)^2 + \beta C_i (\sigma_i^*)^2,
\end{equation}
where we used the property $\|A V^T\|_F = \|A\|_F$ to simplify the third term of Eq.~\ref{eq:act-aware-condi-Substituting}, and $C_i = \|(\delta X U)_i\|_2^2$ measures the empirical energy of the activation error projected onto the $i$-th singular direction. Setting $\partial \mathcal{L}(\sigma_i^*)/\partial \sigma_i^* = 0$ yields:
\begin{equation}
(\sigma_i^* - \sigma_i) + 2\lambda(\sigma_i^* - t) + 2\beta C_i \sigma_i^* = 0.
\end{equation}
Solving for $\sigma_i^*$ gives the closed-form solution:
\begin{equation}
\sigma_i^* = \frac{\sigma_i + 2\lambda t}{1 + 2\lambda + 2\beta C_i}.
\label{eq:sigma_update}
\end{equation}

To preserve the core representational information of pre-trained weights, we introduce a spectral energy preservation strategy. In particular, we identify the dominant singular values whose cumulative squared sum accounts for $\tau$\% of the total spectral energy:
\begin{equation}
    \sum_{j=1}^{p} \sigma_{(j)}^2 / \sum_{j=1}^{r} \sigma_{(j)}^2 \geq \tau,
\end{equation}
where $\sigma_{(j)}$ are sorted in descending order. These dominant singular values are kept immutable during optimization. For the remaining tail singular values, we selectively apply the update rule in Eq.~\ref{eq:sigma_update} only when it would increase their magnitude ($\sigma_i^* > \sigma_i$). This non-monotonic refinement prevents excessive attenuation of weak signals while effectively rectifying ill-conditioned spectral distributions. The updated weights are then represented as $W_{k+1} = Z^* = U \Sigma^* V^T$. This process is repeated iteratively until the maximum number of iterations is reached.

Compared with CondiQuant~\cite{liu2025condiquant}, ACNR offers three key advantages: (1) The activation-aware penalty term in Eq.~\ref{eq:act-aware-condi} introduces a directional suppression via $C_i$ in the denominator of Eq.~\ref{eq:sigma_update}, prioritizing robustness against the actual error distribution of quantized activations. (2) The proximal point algorithm is more stable for low-bit quantization, as it avoids the noisy gradient updates present in PGD. (3) The spectral energy preservation strategy selectively protects dominant singular directions while adaptively adjusting tail singular values, better balancing condition number reduction with weights' representational capacity preservation.

As shown in the green curve of Fig.~\ref{sam_cond_comparison}, the condition numbers of SAM models are reduced significantly, thereby yielding better weight distribution for quantization. Note that ACNR is performed before the quantization process and does not result in additional training complexity and inference overhead, thereby making it hardware-compatible.

\begin{figure}[htbp]
    \centering
    \includegraphics[width=\linewidth]{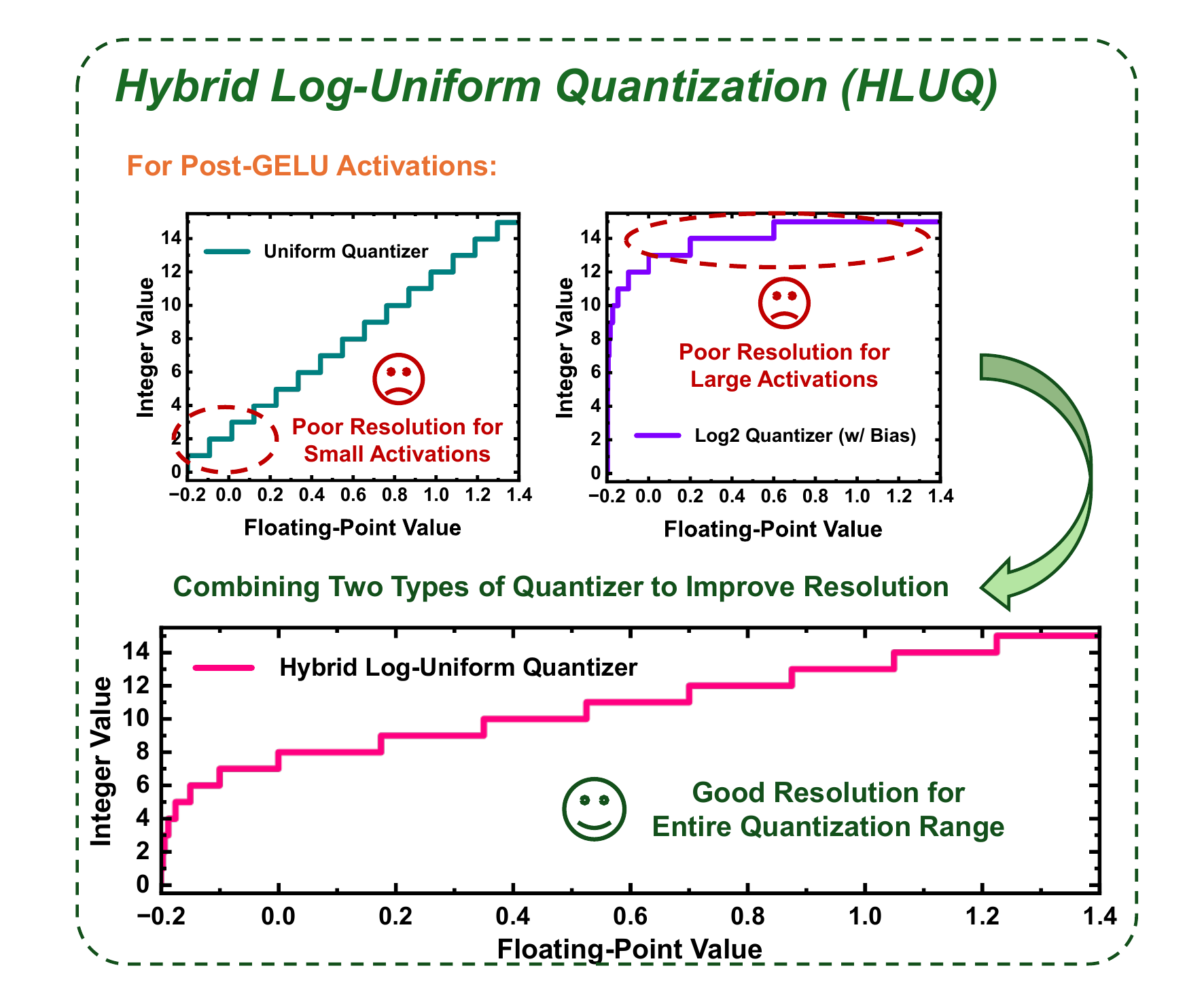}
    \caption{Hybrid Log-Uniform Quantization (HLUQ) applies the power-of-two quantizer for densely clustered small values and the uniform quantizer for sparse but widely distributed large values, effectively handling the skewed and long-tailed post-GELU activations.}
    \label{Fig_HLUQ}
\end{figure}

\subsubsection{Hybrid Log-Uniform Quantization}
\label{Subsubsection_HLUQ}

\textbf{Challenge 2.} The \textbf{second challenge} of quantization arises from the skewed and long-tailed distribution of post-GELU activations. As depicted in Fig.~\ref{Fig_Post_GELU_Acts}, over 90\% of activations are densely concentrated within -0.2 to 0, whereas a sparser but widely distributed subset of large values, critical for inference accuracy, extends from 0 to 0.8. As illustrated in Fig.~\ref{Fig_HLUQ}, existing power-of-two and uniform quantizers fail to effectively handle this distribution. The power-of-two quantizer efficiently captures small, densely clustered values by allocating most of its quantization grids to this range. However, its exponentially spaced grid structure results in insufficient representation density for larger values, leading to high quantization errors in the upper range. On the other hand, the uniform quantizer provides consistent resolution across the full range, making it better suited for large, widely spaced values, but it lacks sufficient resolution for small activations, introducing substantial quantization errors. This problem is further exacerbated as bit-width decreases, particularly in ultra-low-bit settings, where the limited number of grids imposes severe constraints. Although non-uniform quantizers may provide a better alternative, their hardware complexity poses deployment challenges~\cite{Lookup_Table,Data-free,DAQ} and often necessitates substantial QAT-based retraining~\cite{APoT,UPoT}. To effectively address this issue, an efficient, hardware-compatible quantizer that captures both small and large activations tailored for PTQ is required.

\textbf{Solution.} To address the above challenge, we propose Hybrid Log-Uniform Quantization (HLUQ), a method that reduces quantization errors in post-GELU activations by combining power-of-two and uniform quantization. Specifically, HLUQ applies power-of-two quantization to densely packed small values, where fine-grained precision is crucial, and uniform quantization to sparse and long-tailed large values, ensuring minimal quantization error:

\begin{equation}
x_q =
\begin{cases}
\text{clamp} \left( \left\lfloor -\log_{2}\frac{x}{s_{1}} \right\rceil, 0, \hat{b} \right), & \text{if } x \leq s_{1}, \\[0.5em]
\text{clamp} \left( \left\lfloor \frac{x-s_{1}}{s_{2}} \right\rceil, \hat{b}, 2^k - 1 \right), & \text{if } x > s_{1}.
\end{cases}
\label{Eq_Hybrid_Quant}
\end{equation}

\begin{equation}
x \approx \hat{x} =
\begin{cases}
s_{1} \cdot 2^{-x_{q}}, & \text{if } x_q \leq \hat{b}, \\
s_{2} \cdot x_{q} + s_{1}, & \text{if } x_q > \hat{b}.
\end{cases}
\label{Eq_Hybrid_Dequant}
\end{equation}

\noindent Here, $s_1$ defines the power-of-two quantization scale, and $s_2$ determines the uniform quantization scale. The threshold $\hat{b}$ partitions the quantization grid, assigning power-of-two quantization to $[0, \hat{b}]$ and uniform quantization to $[\hat{b}, 2^k - 1]$.

By adjusting $s_1$ and $\hat{b}$, HLUQ offers a high degree of adaptability, allowing it to accommodate various activation distributions. If $s_1$ spans the full range and $\hat{b}$ is close to $2^k - 1$, HLUQ behaves like the power-of-two quantizer. In contrast, if $s_1$ and $\hat{b}$ are near zero, it essentially acts as a uniform quantizer, ensuring uniform step sizes. For skewed and long-tailed post-GELU activations, HLUQ can be configured with intermediate values of $s_1$ and $\hat{b}$, allowing it to capture small, densely distributed values using power-of-two quantization while retaining precision for large, sparse values through uniform quantization, as illustrated in Fig.~\ref{Fig_HLUQ}. Furthermore, HLUQ maintains the hardware efficiency of power-of-two and uniform quantization, introducing only minimal overhead to partition the input at $s_1$.

To initialize $\hat{b}$, $s_1$, and $s_2$, we introduce two auxiliary parameters, $\alpha$ and $\beta$, and search their optimal values during initial calibration by minimizing the following objective function:

\begin{equation}
\mathop{\arg} \mathop{\min}_{\alpha,\beta} \mathbb{E} \left[ \| \mathbf{X}\mathbf{W} - \hat{\mathbf{X}}\mathbf{W} \|_F^2 \right]
\label{Eq_HLUQ_Cali}
\end{equation}

\noindent Here, $\alpha$ partitions the original activation range $r$, defining the quantization scales $s_1$ and $s_2$ as $s_1 = \alpha \cdot r$ and $s_2 = (1 - \alpha) \cdot r$. Meanwhile, $\beta$ determines the allocation of quantization grids between power-of-two and uniform quantization segments, setting the threshold $\hat{b}$ as $\hat{b} = \beta \cdot (2^k - 1)$. Once $\alpha$ and $\beta$ are obtained, the scales $s_1$, $s_2$, and the grid threshold $\hat{b}$ are configured to initialize the HLUQ quantizer for subsequent reconstruction training.

\subsubsection{Channel-Aware Grouping}

\begin{figure}[ht]
    \centering
        \includegraphics[width=\linewidth]{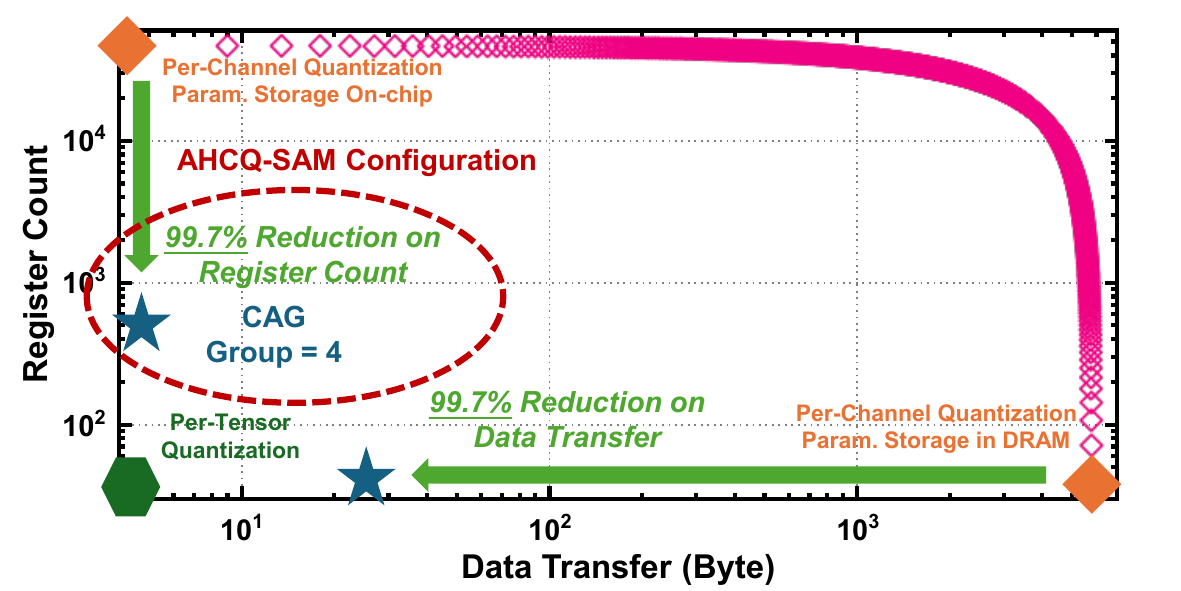}
        \caption{Hardware cost analysis of the linear projection layer in the SAM-H decoder under different quantization granularities.}
        \label{Fig_Hardware_Cost}
\end{figure}

\textbf{Challenge 3.} The \textbf{third challenge} of quantization arises from high inter-channel variation, particularly in the activations of \texttt{Query}/\texttt{Key}/\texttt{Value} linear projections in the attention module and \texttt{Linear} projections in the MLP module. These variations originate from LayerNorm operations and the unique activation distributions of the SAM mask decoder. As shown in Fig.~\ref{Fig_Lin_Proj_Acts}, activation ranges in the Token-to-Image \texttt{Value} linear projection exhibit significant inter-channel disparities, making per-tensor quantization ineffective due to the challenge of finding a single optimal scale factor and zero point~\cite{ACIQ}. A large scale factor, selected to accommodate high-range channels, leads to reduced quantization granularity for low-range channels, often causing them to be rounded to zero. Conversely, a small scale factor, optimized for low-range channels, results in severe clipping in high-range channels, leading to significant information loss~\cite{Dynamic_Dual}. Moreover, while a zero point could compensate for activation skewness, the varying inter-channel median values prevent a single zero point from being optimal for all channels. While per-channel quantization can effectively mitigate quantization errors, it introduces considerable hardware inefficiencies. As depicted in Fig.~\ref{Fig_Hardware_Cost}, transferring per-channel quantization parameters between DRAM and compute units incurs a memory access overhead of up to several kB per layer~\cite{Energy_Problem,BitSplitStitching}. Storing these parameters on-chip can eliminate the transfer cost, but at the expense of a significantly larger memory footprint, requiring tens of thousands of on-chip registers, thereby increasing chip area utilization. To address this trade-off, a granularity-aware quantization approach with hardware co-optimization is essential for balancing quantization accuracy and deployment efficiency.

\begin{figure}[th]
    \centering
        \includegraphics[width=\linewidth]{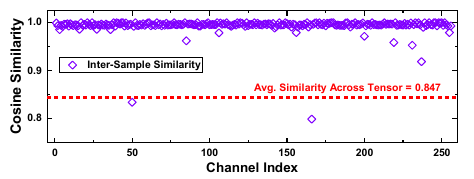}
        \caption{Cosine similarity of normalized quantization parameter across 100 samples for each channel.}
        \label{Fig_Chan_Sim}
\end{figure}

\begin{figure}[htbp]
    \centering
    \includegraphics[width=\linewidth]{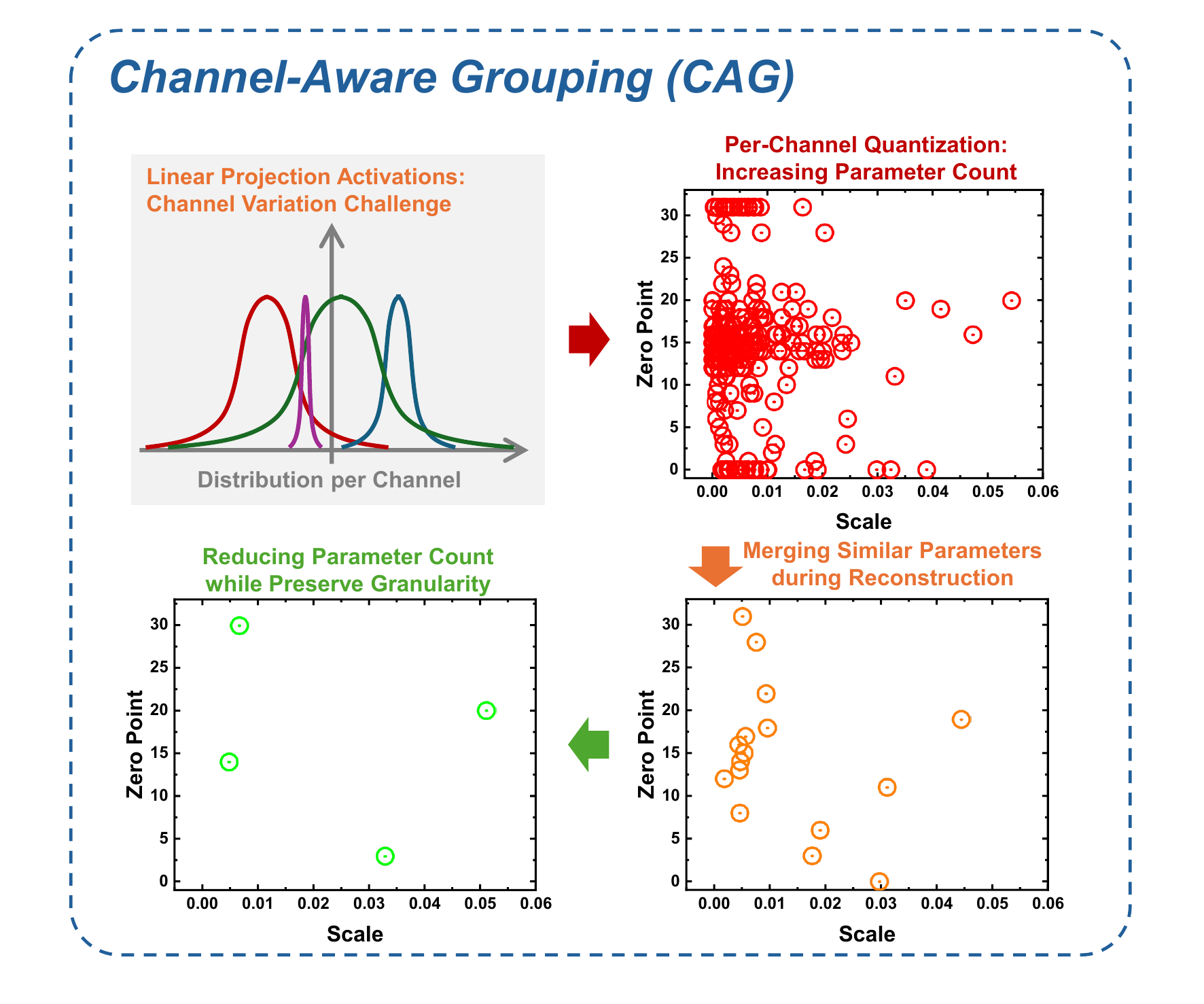}
    \caption{Channel-Aware Grouping (CAG) progressively groups channels with similar activation distributions and assigns them shared quantization parameters, effectively capturing the high inter-channel variation activations.}
    \label{Fig_CAG}
\end{figure}

\textbf{Solution}. To address the above challenge, we first investigate the statistical properties and interestingly reveal that although linear projection activations exhibit substantial inter-channel variation, their characteristics remain consistent across different input samples. As shown in Fig.~\ref{Fig_Chan_Sim}, the cosine similarity of normalized quantization parameters, searched over 100 samples per channel, is consistently close to 1.0, indicating that the optimal quantization parameters for each channel are largely invariant across samples. This stability suggests the feasibility of employing shared quantization parameters within grouped channels to improve hardware efficiency without compromising quantization accuracy. 

\begin{algorithm}[thbp]
\caption{Channel-Aware Grouping}
\label{Algorithm_CAG}
\begin{algorithmic}[1]
\REQUIRE Total iterations $T$, \\ milestones $t \in \{T_1, T_2, \dots, T_J\}$
\ENSURE $K$ groups of optimized parameters ${\{s_i,z_i\}}_{i=1}^{K}$
\STATE Initialize quantization parameters ${\{s_i,z_i\}}_{i=1}^{N}$ for $N$ channels, $G \gets N$
\FOR{$t = 1, 2, ..., T$}
\STATE Update ${\{s_i,z_i\}}_{i=1}^{G}$ according to Eq.~\ref{Eq_Block_Recon}
\IF{$t \in \{T_1, T_2, \dots, T_J\}$}
\STATE Apply K-Means to ${\{s_i, z_i\}}_{i=1}^{G}$ to form centroids ${\{s_i, z_i\}}_{i=1}^{G_{T_{j}}}$
\STATE Assign each channel to the nearest centroid  \\ to form new groups
\STATE Update number of groups $G \gets G_{T_{j}}$
\ENDIF
\ENDFOR
\RETURN ${\{s_i,z_i\}}_{i=1}^{K}$
\end{algorithmic}
\end{algorithm}

Thus, as shwon in Fig.~\ref{Fig_CAG}, we propose Channel-Aware Grouping (CAG), which achieves accuracy comparable to per-channel quantization while significantly reducing parameter overhead, thereby enabling efficient on-chip deployment. Unlike static per-group quantization~\cite{RPTQ}, the fundamental concept of CAG is to progressively group channels with similar activation distributions and assign them shared quantization parameters. As outlined in Algorithm~\ref{Algorithm_CAG}, the process begins with quantization parameter initialization (scales and zero points) for each channel via model calibration, treating each channel as an independent group initially. Next, block-wise reconstruction is applied to refine both quantization parameters and weights, minimizing quantization error as formulated in Eq.~\ref{Eq_Block_Recon}, ensuring alignment with each group's distribution characteristics. At designated milestones, channels with similar quantization parameters are progressively clustered, and the resulting centroids are adopted as shared quantization parameters for all channels within a group. This iterative process continues until the target group count is reached. Fig.~\ref{Fig_CAG} illustrates an example of this grouping process for linear projection activations in Token-to-Image \texttt{Value} cross-attention, demonstrating how CAG effectively reduces the number of groups while preserving quantization performance. With a group number of 4, the edge accelerator can minimize quantization parameter overhead, reducing either data transmission or on-chip storage by 99.7\%, as shown in Fig.~\ref{Fig_Hardware_Cost}. In AHCQ-SAM, we adopt on-chip storage for quantization parameters, requiring only 144 registers to store the scales and zero points of these 4 groups under 4-bit quantization. As depicted in Fig.~\ref{Fig_CAG_Dependence}, CAG maintains accuracy comparable to per-channel quantization, offering high hardware efficiency while significantly enhancing model performance.

\subsubsection{Logarithmic Nonlinear Quantization}

\begin{figure}[htbp]
    \centering
    \includegraphics[width=\linewidth]{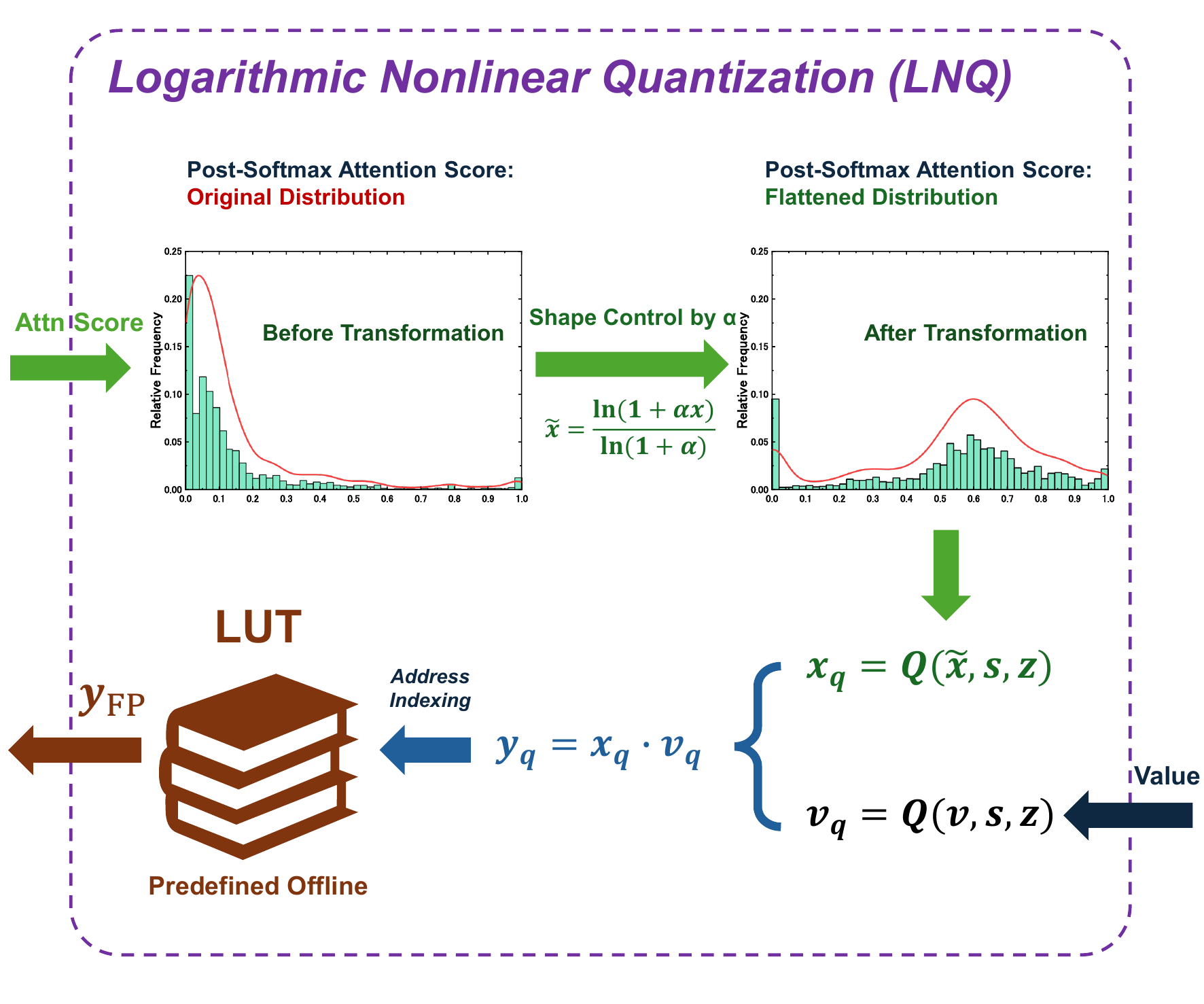}
    \caption{Logarithmic Nonlinear Quantization (LNQ) employs logarithmic transformations to adaptively adjust quantization resolution for exponentially scaled and heterogeneous attention scores}
    \label{Fig_LNQ}
\end{figure}

\textbf{Challenge 4.} The \textbf{fourth challenge} of quantization arises from the exponentially scaled range and varied distribution patterns of post-Softmax attention scores~\cite{PTQ4SAM}. As illustrated in Fig.~\ref{Fig_Softmax_Dist}, these scores reach as low as $10^{-15}$, which far exceeds the representational capacity of standard power-of-two quantizers. For instance, a 4-bit power-of-two quantizer can only represent a minimum value of $2^{-16} \approx 10^{-5}$ and would inevitably round the majority of small attention scores to zero, causing significant information loss. The uniform quantizer also suffers from insufficient resolution for small attention scores. In addition, the attention scores exhibit distinct patterns across different layers. While Token-to-Image Attention scores present a relatively broader and uniform distribution spanning from $10^{-15}$ to $10^{-2}$, Self Attention scores are densely clustered near $10^{-5}$. In contrast, Image-to-Token Attention scores show a skewed distribution that reaches toward $1.0$. This diversity underscores the need for an adaptive quantizer based on flexible transformations that can effectively accommodate such wide ranges and shifting patterns.

\textbf{Solution.} Motivated by the above challenge, as shown in Fig.~\ref{Fig_LNQ}, we propose a Logarithmic Nonlinear Quantization (LNQ), which employs a logarithmic transformation operator $\mathcal{T}$ to reshape the distribution of attention scores $x$:
\begin{equation}
    \tilde{x} = \mathcal{T}(x; \alpha) = \frac{\ln(1 + \alpha x )}{\ln(1 + \alpha)},
\end{equation}
where $\alpha > 0$ represents the shape factor that governs the intensity of the nonlinear mapping. The transformed $\tilde{x}$ are subsequently mapped to a $b$-bit integer representation $x_q$ through a standard uniform quantizer as presented in Eq.~\ref{Eq_Uniform_Quant}. To reconstruct the floating-point approximation $\hat{x}$, the full-precision input $x$ is approximated as follows:
\begin{equation}
 x \approx \hat{x} = \mathcal{T}^{-1} \left( s \cdot (x_{q} - z); \alpha \right) = \frac{(1 + \alpha)^{s \cdot (x_{q} - z)} - 1}{\alpha}.
\end{equation}

The operator $\mathcal{T}(\cdot)$ adaptively adjusts the input distribution as a function of the shape factor $\alpha$. When $\alpha \to 0$, the operator converges to an identity mapping, thereby preserving the original distribution when nonlinear warping is not required. As $\alpha$ increases, the operator presents flattened curvature, effectively compressing sparse high-magnitude outliers while logarithmically expanding the dense low-magnitude regions. Thereby, this redistribution reallocates a higher quantization resolution to the infinitesimal scores.

The $\alpha$ is initialized via a grid search. We define a candidate set $\mathcal{S}_{\alpha}$ and determine the optimal initial value $\alpha_{init}$ by minimizing the $L_2$ reconstruction error:

\begin{equation}
   \alpha_{init} = \mathop{\arg\min}_{\alpha \in \mathcal{S}_{\alpha}} \mathcal{L}_{L_2}(x, \hat{x}).
\end{equation}
By assigning an independent $\alpha$ to each attention layer, the LNQ adaptively applies the optimal transformation intensity tailored to disparate attention patterns. From a deployment perspective, the nonlinear transformation and subsequent uniform quantization can be fused into a static threshold table $\bm{T}_{th}$ during the offline phase. In practice, the inference process involves performing matrix multiplication between the quantized post-Softmax scores and \texttt{Value} $\bm{V}$ to obtain integer intermediate results. These intermediate results then serve as indices to access a precomputed dequantization Look-up Table (LUT), which restores the output for subsequent operations. By repurposing redundant on-chip BRAM as LUTs, we eliminate the computational overhead of logarithmic and exponential functions, ensuring high hardware efficiency.

\subsection{Hardware Architecture Co-optimization}
\label{Subsection_Hardware_Architecture_Co-optimization}

\begin{figure}[thbp]
    \centering
        \includegraphics[width=\linewidth]{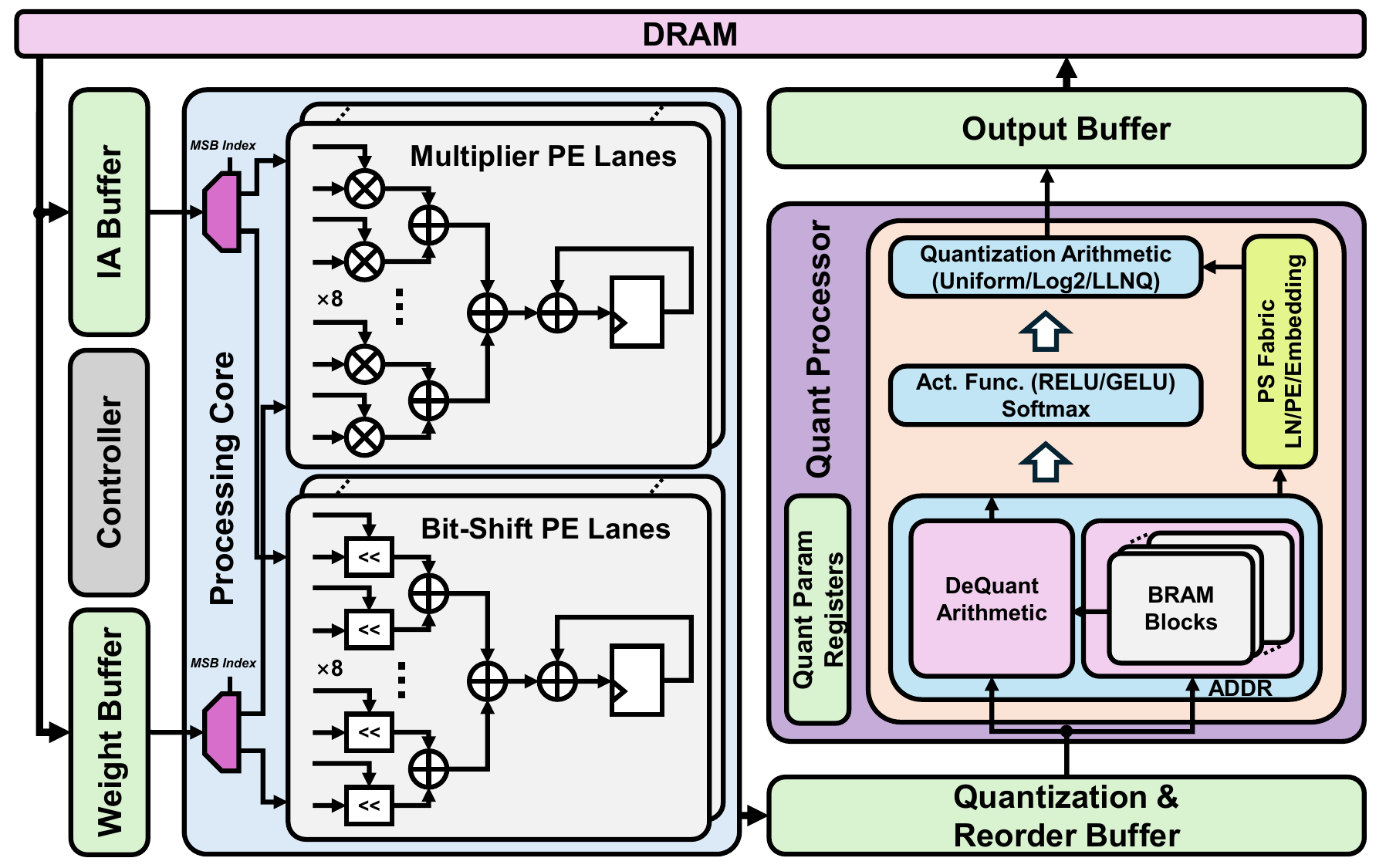}
        \caption{Hardware–algorithm co-designed FPGA accelerator architecture enabling efficient AHCQ-SAM implementation.}
        \label{Fig_Hardware_Architecture}
\end{figure}

To fully leverage the proposed AHCQ-SAM quantization scheme, we co-design an FPGA-based hardware accelerator to bridge algorithm–hardware efficiency. The accelerator is deployed on a Xilinx Zynq UltraScale+ MPSoC FPGA, and the overall architecture is illustrated in Fig.~\ref{Fig_Hardware_Architecture}. Off-chip DDR4 memory buffers intermediate activations, while four groups of on-chip BRAM-based buffers are utilized for activation storage. The design supports two processing element (PE) configurations: \textbf{(1)} an 8-input integer multiplier–accumulator PE for uniform quantization, where integer inputs and outputs are processed with partial sums accumulated in an integer accumulator; and \textbf{(2)} an 8-input decimal bit-shift–accumulator PE for power-of-two quantization, which accepts integer inputs and produces decimal outputs using a decimal accumulator. Dequantization, activation function, and quantization operations are deeply pipelined in the programmable logic (PL), whereas floating-point computations, including normalization, positional encoding, and embedding, are executed in the processing system (PS) to simplify hardware design.

\textbf{HLUQ:} Matrix multiplication within the HLUQ quantizer is executed using both multiplier PE lanes and bit-shift PE lanes. The uniform quantization branch is mapped to the multiplier PE lanes, while the power-of-two quantization branch is mapped to the bit-shift PE lanes. The grid ratio between the two quantizers follows the $2^{-n}$ rule, such that the most significant bits (MSBs) of quantized values act as label bits. These label bits enable efficient routing of quantized values to the corresponding PEs. The outputs of both branches are subsequently fused during dequantization in the quantization processor to generate the final result.

\textbf{CAG:} Quantization parameters are stored in a small set of on-chip registers, and a dedicated quantization processor is implemented in the PL fabric. After linear projection, weights and activations are reordered according to channel indices within a reorder buffer, ensuring sequential alignment of grouped channels. Weights are preprocessed offline, while activations are reordered on-chip. By adopting counter-based logic for parameter switching, the design reduces hardware complexity without sacrificing computational efficiency.

\textbf{LNQ:} Redundant BRAM resources are repurposed as LUTs to support dequantization in LNQ. Following the Softmax operation, integer post-activations are used as addresses to access precomputed results stored in BRAM, which are subsequently processed by the dequantization unit to recover FP values. For 4-bit quantization, all possible post-activation values produced by PTQ are enumerated offline, leading to a BRAM page size smaller than a 10-bit address space. The resulting memory overhead is negligible, as BRAM resources are typically underutilized in FPGA accelerator designs.

\section{Experimentation}

\subsection{Experiment and Implementation Details}

\subsubsection{Model and Datasets}

Regarding SAM, to evaluate the effectiveness of AHCQ-SAM, we conduct instance segmentation experiments on the COCO dataset~\cite{COCO} using the mean Average Precision (mAP) metric. The selected detectors include CNN-based Faster RCNN~\cite{FasterRCNN} and YOLOX~\cite{Yolox}, as well as Transformer-based H-Deformable-DETR~\cite{HDETR} and DINO~\cite{DINO}. The bounding boxes generated by these detectors serve as prompts for SAM. For CNN-based detectors, the box threshold is set to 0.05, while Transformer-based detectors utilize a set of 100 adaptive anchors. Regarding SAM2, we evaluate the performance on the Video Object Segmentation (VOS) task using the standard $\mathcal{J}$\&$\mathcal{F}$ metric across the DAVIS~\cite{pont20172017,caelles20192019} and Segment Anything Video (SA-V)~\cite{ravisam} datasets.

In SAM, following the PTQ4SAM framework~\cite{PTQ4SAM}, we randomly sample 32 training images for both model calibration and block-wise optimization. In SAM2, we randomly sample 8 training videos, with the first 4 frame clips utilized for calibration and optimization. For both SAM and SAM2, we use RMSE to calibrate the weight quantization parameters, while the MinMax approach is used for initialization of activations.

\subsubsection{Implementation Details}

For both SAM and SAM2, block reconstruction is performed over 20,000 iterations on the block. To ensure fair comparisons~\cite{PTQ4SAM,Pd-quant,Qdrop,Ptq4vit}, we exclude the first and last layers or blocks from quantization while keeping all others quantized. The learning rate is initialized at $4 \times 10^{-5}$ and decayed via a cosine annealing scheduler. The batch size is set to 1.

ACNR is selectively applied to weight matrices with a condition number exceeding 100, utilizing a maximum of 200 iterations. The $\lambda$, $\beta$, and $\tau$ are empirically set to 0.003, 0.001, and 80, respectively. HLUQ is applied to all Linear-2 activations in MLP blocks, with initial scales and grid thresholds determined by searching $\alpha \in \{0.1,0.3,0.5\}$ and $\beta \in \{\frac{1}{2},\frac{1}{4},\frac{1}{8}\}$. CAG is applied to all linear projection activations in \texttt{Query}/\texttt{Key}/\texttt{Value} projections of attention blocks and Linear-1 activations in MLP blocks, using a group number of 4. For other activations, we adopt per-tensor asymmetric quantization. For weights, we apply per-channel asymmetric quantization to maintain alignment with baseline settings. LNQ is applied to all post-Softmax scores in SAM, and post-Softmax scores for the image encoder in SAM2. The candidate set $\mathcal{S}_{\alpha} = \{1, 10, 50, 100, 200, 500\}$.

\begin{table*}[t]
\centering
\caption{Quantization performance of AHCQ-SAM on the SAM series for instance segmentation (COCO dataset). Reference floating-point (FP32) mAP values for SAM-B, SAM-L, and SAM-H are provided for comparative analysis. AHCQ-SAM achieves a new state-of-the-art performance, outperforming all existing baselines. \textbf{Note:} This paper reports the corrected performance of PTQ4SAM and QDrop after fixing a bug (refer to code repository.) in the PTQ4SAM~\cite{PTQ4SAM} framework.}
\label{Table_Exp_Result}
\begin{tabular}{c|c|c|ccc|c|ccc|c|ccc}
\toprule
\multirow{2}{*}{\textbf{Detector}} & \multirow{2}{*}{\textbf{Method}} & \multicolumn{4}{c|}{\textbf{SAM-B}} & \multicolumn{4}{c|}{\textbf{SAM-L}} & \multicolumn{4}{c}{\textbf{SAM-H}} \\
\cmidrule(lr){3-6} \cmidrule(lr){7-10} \cmidrule(lr){11-14}
 & & \textbf{FP} & \textbf{W4A4} & \textbf{W5A5} & \textbf{W6A6} & \textbf{FP} & \textbf{W4A4} & \textbf{W5A5} & \textbf{W6A6} & \textbf{FP} & \textbf{W4A4} & \textbf{W5A5} & \textbf{W6A6} \\
\midrule
\multirow{5}{*}{\textbf{Faster R-CNN}} 
& BRECQ & \multirow{5}{*}{33.1} & 0.2 & 16.7 & 28.0 & \multirow{5}{*}{36.0} & 5.0 & 31.8 & 35.2 & \multirow{5}{*}{36.8} & 17.5 & 31.3 & 35.8 \\
& QDrop & & 2.3 & 12.9 & 26.2 & & 0.8 & 31.9 & 35.0 & & 6.0 & 32.6 & 36.0 \\
& PTQ4SAM & & 2.7 & 14.4 & 26.8 & & 2.4 & 33.0 & 35.5 & & 6.7 & 33.3 & 36.2 \\
% & AHCPTQ & & 11.7 & 27.5 & 31.6 & & 27.4 & 34.8 & 35.6 & & 31.4 & 35.7 & 36.3 \\
& AHCQ-SAM & & \textbf{17.9} & \textbf{29.2} & \textbf{32.0} & & \textbf{29.5} & \textbf{35.1} & \textbf{35.7} & & \textbf{32.6} & \textbf{36.1} & \textbf{36.4} \\
\midrule
\multirow{5}{*}{\textbf{YOLOX}}
& BRECQ & \multirow{5}{*}{37.2} & 0.2 & 19.0 & 31.9 & \multirow{5}{*}{40.4} & 6.3 & 35.3 & 39.4 & \multirow{5}{*}{41.0} & 19.7 & 34.7 & 39.7 \\
& QDrop & & 2.6 & 15.6 & 30.3 & & 1.0 & 36.2 & 39.4 & & 6.8 & 36.0 & 40.1 \\
& PTQ4SAM & & 3.8 & 18.4 & 30.9 & & 2.4 & 37.1 & 39.9 & & 7.4 & 37.1 & 40.3 \\
% & AHCPTQ & & 13.4 & 31.8 & 35.4 & & 31.0 & 39.1 & 40.0 & & 35.2 & 40.0 & 40.4 \\
& TODO & & \textbf{20.9} & \textbf{32.3} & \textbf{35.6} & & \textbf{33.5} & \textbf{39.4} & \textbf{40.1} & & \textbf{35.9} & \textbf{40.3} & \textbf{40.5} \\
\midrule
\multirow{5}{*}{\textbf{H-DETR}} 
& BRECQ & \multirow{5}{*}{38.2} & 0.3 & 11.2 & 32.0 & \multirow{5}{*}{41.5} & 5.2 & 36.1 & 40.4 & \multirow{5}{*}{42.0} & 19.1 & 35.3 & 40.6 \\
& QDrop & & 2.0 & 13.1 & 30.5 & & 1.3 & 37.0 & 40.3 & & 6.9 & 37.0 & 41.1 \\
& PTQ4SAM & & 2.8 & 16.9 & 30.7 & & 2.6 & 38.1 & 40.9 & & 7.1 & 38.0 & 41.4 \\
% & AHCPTQ & & 14.1 & 32.1 & 36.6 & & 32.3 & 40.3 & 41.0 & & 35.6 & 40.9 & 41.5 \\
& AHCQ-SAM  & & \textbf{20.7} & \textbf{33.7} & \textbf{37.2} & & \textbf{34.0} & \textbf{40.4} & \textbf{41.1} & & \textbf{37.0} & \textbf{41.2} & \textbf{41.6} \\
\midrule
\multirow{5}{*}{\textbf{DINO}} 
& BRECQ & \multirow{5}{*}{44.5} & 0.2 & 13.5 & 34.8 & \multirow{5}{*}{48.6} & 3.6 & 41.4 & 47.0 & \multirow{5}{*}{49.1} & 20.7 & 40.3 & 47.2 \\
& QDrop & & 1.9 & 13.4 & 34.5 & & 1.0 & 42.7 & 47.0 & & 7.0 & 42.4 & 47.9 \\
& PTQ4SAM & & 1.9 & 17.6 & 35.1 & & 2.3 & 44.1 & 47.8 & & 8.9 & 43.8 & 48.2 \\
% & AHCPTQ & & 16.8 & 36.7 & 41.9 & & 36.6 & 46.8 & 47.9 & & 41.2 & 47.6 & 48.3 \\
& AHCQ-SAM  & & \textbf{21.2} & \textbf{38.0} & \textbf{42.7} & & \textbf{40.2} & \textbf{47.2} & \textbf{48.0} & & \textbf{42.4} & \textbf{47.9} & \textbf{48.4} \\
\bottomrule
\end{tabular}
\end{table*}

\begin{table*}[t]
\centering
\caption{Quantization performance on SAM2 for Video Object Segmentation (VOS). The results on $\mathcal{J}$\&$\mathcal{F}$ scores across DAVIS, SA-V Val, and SA-V Test datasets are reported. AHCQ-SAM outperforms all existing baselines. 
}
\label{Table_Exp_Result-sam2}
\begin{tabular}{c|c|c|cc|c|cc|c|cc}
\toprule
\multirow{2}{*}{\textbf{Model}} & \multirow{2}{*}{\textbf{Method}} & \multicolumn{3}{c|}{\textbf{DAVIS}} & \multicolumn{3}{c|}{\textbf{SA-V Val}} & \multicolumn{3}{c}{\textbf{SA-V Test}} \\
\cmidrule(lr){3-5} \cmidrule(lr){6-8} \cmidrule(lr){9-11}
 & & \textbf{FP} & \textbf{W4A4} & \textbf{W6A6} & \textbf{FP} & \textbf{W4A4} & \textbf{W6A6} & \textbf{FP} & \textbf{W4A4} & \textbf{W6A6} \\
\midrule
\multirow{4}{*}{\textbf{SAM2-Tiny}} 
& BRECQ & \multirow{4}{*}{87.96} & 59.92 & 86.20 & \multirow{4}{*}{74.48} & 29.13 & 50.74 & \multirow{4}{*}{77.13} & 28.57 & 53.33 \\
& QDrop & & 65.74 & 86.18 & & 33.96 & 62.34 & & 38.49 & 65.16 \\
& PTQ4SAM & & 63.09 & 85.94 & & 34.79 & 70.65 & & 35.70 & 73.14 \\
& TODO & & \textbf{71.94} & \textbf{86.82} & & \textbf{47.34} & \textbf{72.12} & & \textbf{49.71} & \textbf{73.23} \\
\midrule
\multirow{4}{*}{\textbf{SAM2-Small}}
& BRECQ & \multirow{4}{*}{88.28} & 66.97 & 85.77 & \multirow{4}{*}{76.10} & 32.48 & 45.36 & \multirow{5}{*}{77.43} & 31.33 & 47.16 \\
& QDrop & & 74.07 & 85.94 & & 45.35 & 63.82 & & 46.47 & 66.50 \\
& PTQ4SAM & & 76.67 & 86.04 & & 47.84 & 70.45 & & 50.07 & 73.41 \\
& TODO & & \textbf{78.02} & \textbf{86.32} & & \textbf{48.72} & \textbf{70.85} & & \textbf{52.11} & \textbf{74.46} \\
\midrule
\multirow{4}{*}{\textbf{SAM2-Base+}} 
& BRECQ & \multirow{4}{*}{88.61} & 25.37 & 80.59 & \multirow{4}{*}{76.35} & 20.71 & 47.43 & \multirow{4}{*}{78.74} & 18.48 & 47.24 \\
& QDrop & & 30.59 & 83.48 & & 13.21 & 55.12 & & 14.08 & 57.95 \\
& PTQ4SAM & & 29.19 & 83.29 & & 25.88 & 64.03 & & 25.29 & 65.50 \\
& TODO & & \textbf{31.44} & \textbf{83.69} & & \textbf{26.92} & \textbf{64.69} & & \textbf{26.11} & \textbf{65.95} \\
\bottomrule
\end{tabular}
\end{table*}

\subsection{Experimental Results}

\subsubsection{Results on SAM}

We evaluate AHCQ-SAM against baselines such as BRECQ~\cite{Brecq}, QDrop~\cite{Qdrop}, and PTQ4SAM~\cite{PTQ4SAM}, with results summarized in Tab.~\ref{Table_Exp_Result} across four types of detectors. We did not compare with SAQ-SAM~\cite{zhang2025saq} because it contains a code bug (please see their code repo.). We also emphasize that our AHCQ-SAM complements the SAQ-SAM. For SAM-L and SAM-H models, existing methods suffer from significant accuracy degradation in the 5-bit configuration. For instance, while PTQ4SAM incurs a 3.5\% mAP drop for 5-bit SAM-H with Faster R-CNN, AHCQ-SAM narrows this gap to a mere 0.7\%. Similarly, for 5-bit SAM-H using YOLOX, AHCQ-SAM limits the performance loss to 0.7\%, significantly outperforming the 3.9\% drop observed in PTQ4SAM. The superiority of AHCQ-SAM is even more pronounced in the 4-bit configuration, where baselines struggle to maintain functional utility. Notably, with H-DETR as the detector, AHCQ-SAM restores the mAP of 4-bit SAM-L and SAM-H from a 2.6\% and 7.1\% to a robust 34.0\% and 37.0\%, respectively. Similarly, when paired with DINO, AHCQ-SAM elevates the mAP for 4-bit SAM-L and SAM-H from 2.3\% and 8.9\% to 40.2\% and 42.4\%, respectively. These results clearly demonstrate that AHCQ-SAM surpasses all baselines by a substantial margin, particularly in low-bit configurations. In the challenging SAM-B, AHCQ-SAM consistently achieves superior results. It doubles the mAP compared to PTQ4SAM in the 5-bit, restricts the loss to approximately 1\% in 6-bit, and recovers mAP to a usable level in 4-bit. Specifically, for SAM-B with H-DETR, AHCQ-SAM improves mAP from 2.8\%, 16.9\%, and 30.7\% to 20.7\%, 33.7\%, and 37.2\% for 4-bit, 5-bit, and 6-bit configurations, respectively. Similar performance gains are observed using the DINO detector, where AHCQ-SAM increases the mAP from 1.9\%, 17.6\%, and 35.1\% to 21.2\%, 38.0\%, and 42.7\% for 4-bit, 5-bit, and 6-bit configurations, respectively. These results indicate that AHCQ-SAM consistently surpasses baselines, addressing the challenges identified in previous studies and advancing SAM quantization to lower bit-widths with superior effectiveness.

\subsubsection{Results on SAM2}

To further validate the effectiveness of AHCQ-SAM, we extend our evaluation to the SAM2 family across various model scales, as summarized in Tab.~\ref{Table_Exp_Result-sam2}. AHCQ-SAM consistently outperforms all baseline methods across the DAVIS and SA-V (Val and Test split) datasets. For the lightweight SAM2-Tiny, AHCQ-SAM achieves a remarkable $\mathcal{J}$\&$\mathcal{F}$ score of 71.94\% in the 4-bit configuration on DAVIS, surpassing QDrop and PTQ4SAM by 6.20\% and 8.85\%, respectively. Notably, on the more challenging SA-V Val and SA-V Test sets, AHCQ-SAM respectively elevates the performance of 4-bit SAM2-Tiny to 47.34\% and 49.71\%, representing a substantial improvement of 12.55\% and 11.22\% compared to the best-performing baseline. The advantages of AHCQ-SAM remain consistent as the model capacity increases. For SAM2-Small, AHCQ-SAM maintains high fidelity in 6-bit and restores accuracy in 4-bit, outperforming PTQ4SAM across all datasets with $\mathcal{J}$\&$\mathcal{F}$ scores of 78.02\%, 48.72\%, and 52.11\% on DAVIS, SA-V Val, and SA-V Test, respectively. Even for SAM2-Base+, which poses difficulty for quantization, AHCQ-SAM still maintains superior stability. For example, on 4-bit SAM2-Base+, AHCQ-SAM consistently delivers the highest $\mathcal{J}$\&$\mathcal{F}$ scores by achieving 31.44\%, 26.92\%, and 26.11\% on DAVIS, SA-V Val, and SA-V Test datasets. These results demonstrate that AHCQ-SAM serves as a strong PTQ baseline for SAM2 in video-based tasks.

\subsection{Ablation Studies}

\subsubsection{Ablation of Components}

Tab.~\ref{Table_Framework_Ablation} presents the ablation study to evaluate the contributions of ACNR, HLUQ, CAG, and LNQ within the AHCQ-SAM framework. Using the 5-bit configuration with Faster R-CNN as the detector, the results demonstrate that each component provides a consistent performance uplift across all SAM variants. For instance, for the SAM-B model, incorporating CAG alone yields a significant mAP gain of 8.1\%, while ACNR independently improves the mAP by 4.2\%. For the larger SAM-H model, HLUQ and CAG prove particularly effective, enhancing the mAP by 2.4\% and 2.8\%, respectively. The synergistic effect of these components is most evident when they are integrated. Compared to the baseline, the full AHCQ-SAM configuration restores the mAP of SAM-B, SAM-L, and SAM-H by 8.9\%, 1.1\%, and 3.8\%, reaching 29.2\%, 35.1\%, and 36.1\%, respectively. These findings underscore that while each component targets a specific quantization challenge, their integration yields a collective performance gain.

\begin{table}[htbp]
\centering
\caption{Ablation study of individual components in AHCQ-SAM. The results are evaluated on 5-bit SAM-B/L/H models with Faster R-CNN.}
\label{Table_Framework_Ablation}
\begin{tabular}{cccc|ccc}
\toprule
\textbf{CAG} & \textbf{HLUQ} & \textbf{LNQ} & \textbf{ACNR} & \textbf{SAM-B} & \textbf{SAM-L} & \textbf{SAM-H} \\
\midrule
 & & & & 20.3 & 34.0 & 32.3 \\
$\checkmark$ & & & & 28.4 & 34.9 & 35.1 \\
 & $\checkmark$ & & & 21.2 & 34.5 & 34.7 \\
 & & $\checkmark$ & & 20.5 & 34.1 & 32.4 \\
 & & & $\checkmark$ & 24.5 & 34.2 & 33.1 \\
$\checkmark$ & $\checkmark$ & $\checkmark$ & $\checkmark$ & \textbf{29.2} & \textbf{35.1} & \textbf{36.1} \\
\bottomrule
\end{tabular}
\end{table}

\subsubsection{Ablation of Quantizers}

\begin{figure}[htbp]
\centering
\begin{subfigure}{\linewidth}
    \includegraphics[width=\linewidth]{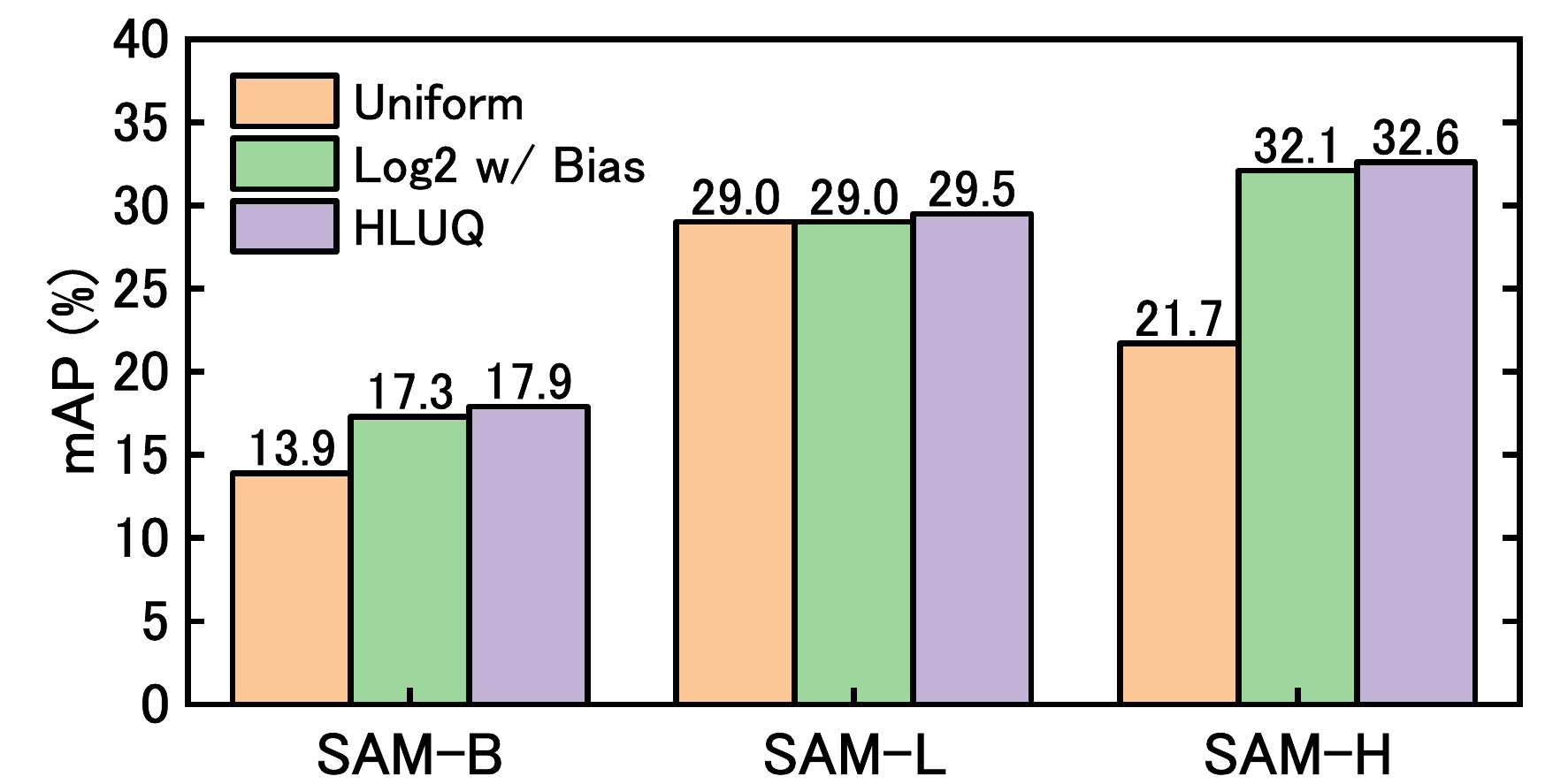}
    \caption{Quantizer for post-GELU Activations}
    \label{Fig_Quantizer_Ablation_PostGELU}
\end{subfigure}

\begin{subfigure}{\linewidth}
    \includegraphics[width=\linewidth]{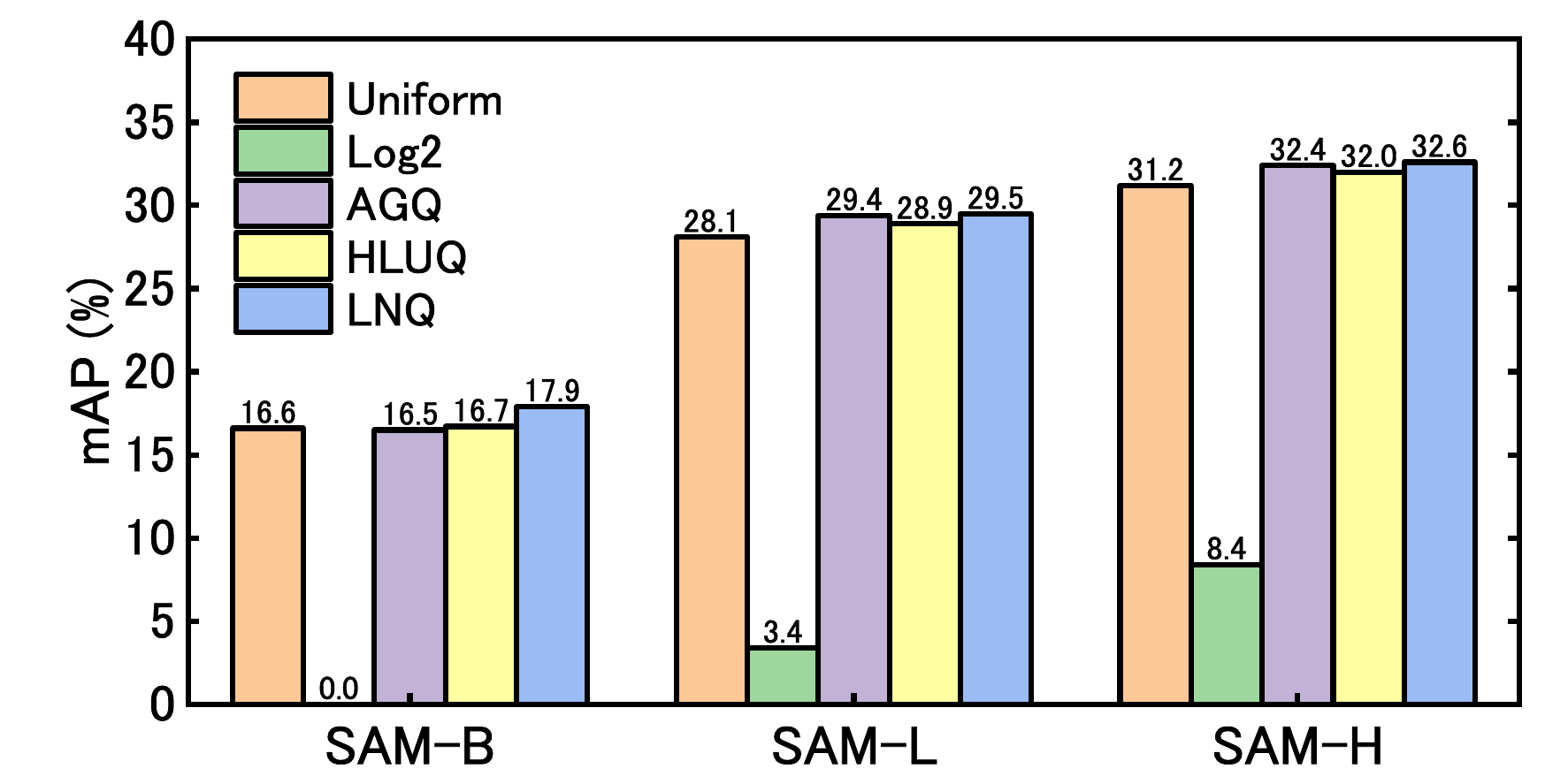}
    \caption{Quantizer for post-Softmax Activations}
    \label{Fig_Quantizer_Ablation_PostSofmax}
\end{subfigure}
\caption{Ablation study of applying different quantizers on 4-bit SAM-B with Faster R-CNN.}
\label{Fig_Quantizer_Ablation}
\end{figure}

To evaluate the effectiveness of the proposed quantizer, we analyze the performance of 4-bit SAM variants by comparing HLUQ and LNQ against various baseline quantizers. Specifically, Fig.~\ref{Fig_Quantizer_Ablation_PostGELU} illustrates the results for post-GELU activations. Since standard power-of-two quantization cannot accommodate the negative values inherent in GELU, a floating-point bias is introduced for a fair comparison. Across all models, HLUQ consistently outperforms both uniform and biased power-of-two quantizers. Notably, for SAM-H, HLUQ achieves a 32.6\% mAP, providing a significant improvement over the 21.7\% uniform baseline and also outperforming the biased power-of-two quantizer, demonstrating its superior capability in accommodating the skewed and long-tailed post-GELU distributions. Furthermore, Fig.~\ref{Fig_Quantizer_Ablation_PostSofmax} presents the ablation results for post-Softmax activations. The standard power-of-two quantizer struggles with the exponentially scaled range of attention scores, failing to extend the value range down to $10^{-15}$ as shown in Fig.~\ref{Fig_Softmax_Dist} and resulting in collapsed performance for SAM-B/L/H. The other quantizers, such as uniform, AGQ~\cite{PTQ4SAM}, and HLUQ may yield competitive results in specific cases, but they exhibit instability across different model sizes. In contrast, LNQ provides a stable and substantial performance recovery, outperforming all alternative quantizers. Specifically, LNQ consistently achieves the highest mAP across all models, reaching 17.9\%, 29.5\%, and 32.6\% for SAM-B, SAM-L, and SAM-H, respectively. These results underscore the superiority of LNQ in managing the exponentially scaled and heterogeneous post-Softmax attention scores.

\subsubsection{Dependence on Group Number}

\begin{figure}[thbp]
    \centering
        \includegraphics[width=\linewidth]{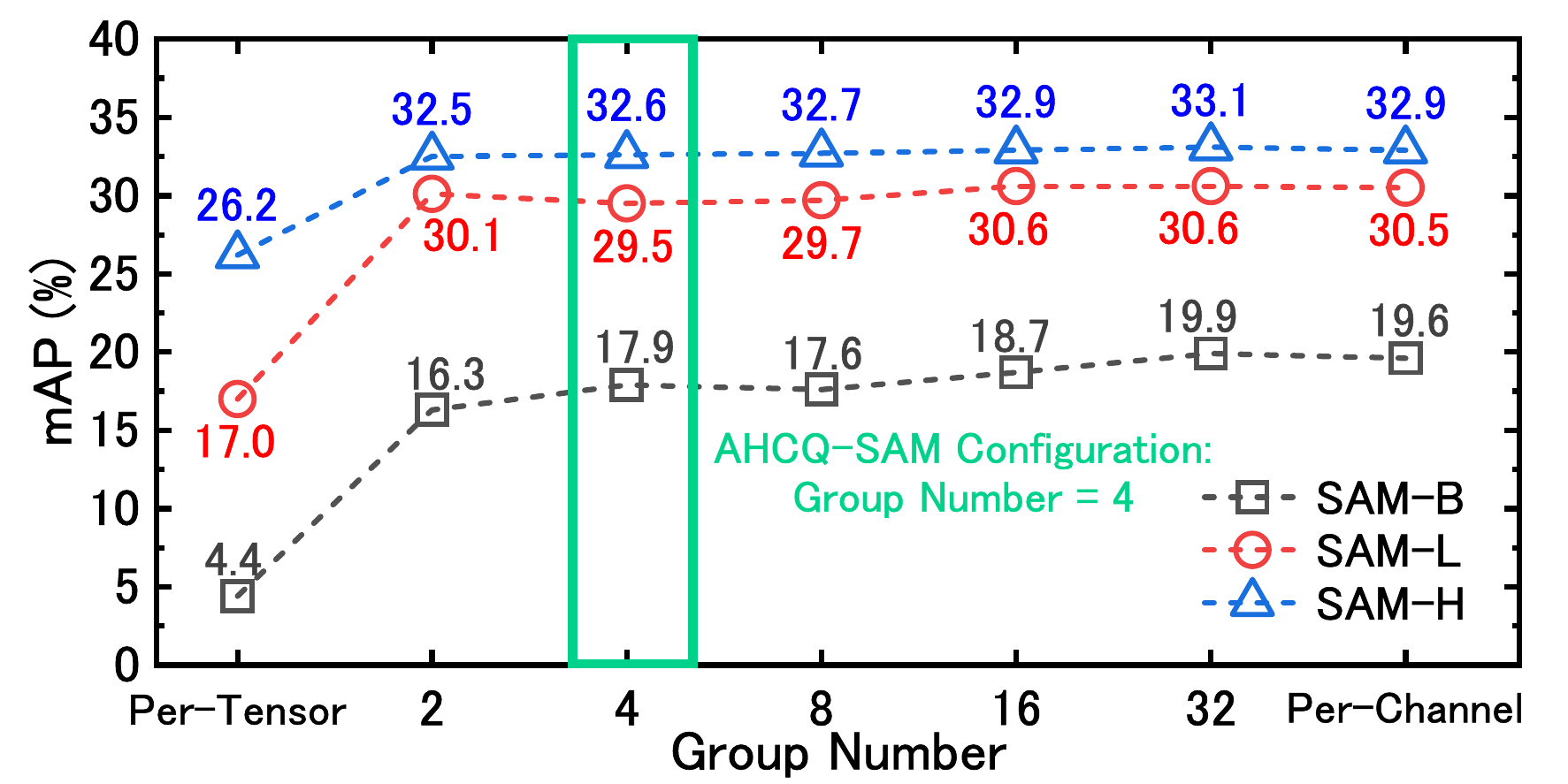}
        \caption{Ablation study of varying group number in CAG for SAM with Faster R-CNN.}
        \label{Fig_CAG_Dependence}
\end{figure}

To investigate the impact of group number in CAG, we evaluate 4-bit SAM variants with Faster R-CNN by varying the number of groups from 2 to 32. The results, illustrated in Fig.~\ref{Fig_CAG_Dependence}, also include per-tensor and per-channel configurations for baseline comparison. For SAM-L and SAM-H, the mAP increases sharply from the per-tensor baseline and begins to saturate at a group count of 2, with only marginal improvements observed as the group number approaches the per-channel limit. In contrast, the SAM-B model exhibits a more gradual recovery, with its performance inflection point occurring at a group count of 4. At this setting, SAM-B maintains a performance gap of approximately 1.7\% compared to the per-channel configuration. To balance cross-model consistency with hardware efficiency, we adopt a group count of 4 as the default configuration in the AHCQ-SAM framework.

\begin{figure*}[htbp]
    \centering
    \includegraphics[width=\textwidth]{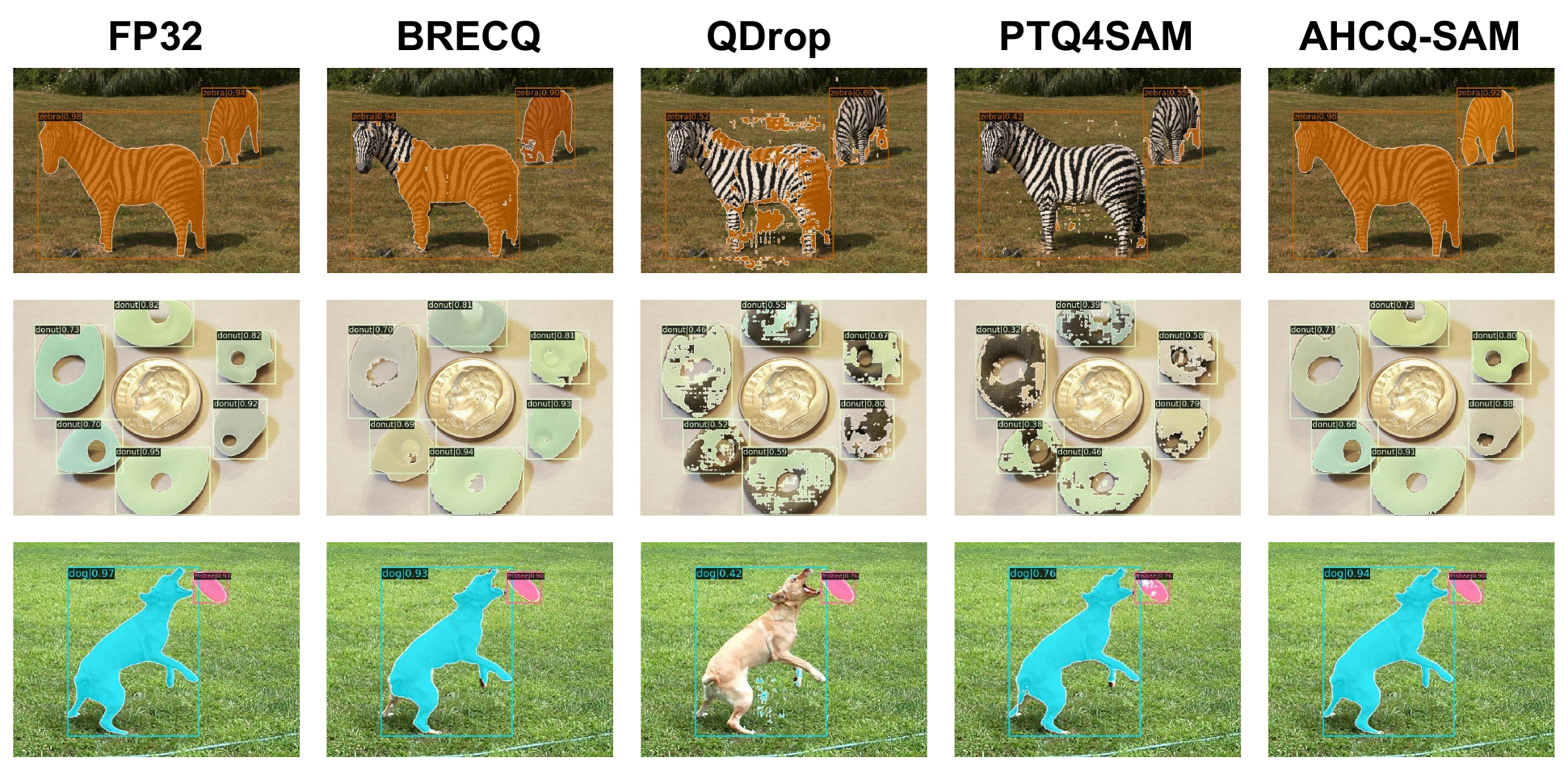}
    \caption{Qualitative comparison of segmentation masks generated by different quantization methods on W4A4 SAM-H with YOLOX. AHCQ-SAM closely matches the floating-point reference, significantly outperforming other baselines.}
    \label{Fig_Visualization}
\end{figure*}

\subsubsection{Comparsion between ACNR and CondiQuant}
\label{sec:Comparsion between ACNR and CondiQuant}

\begin{figure}[thbp]
    \centering
        \includegraphics[width=\linewidth]{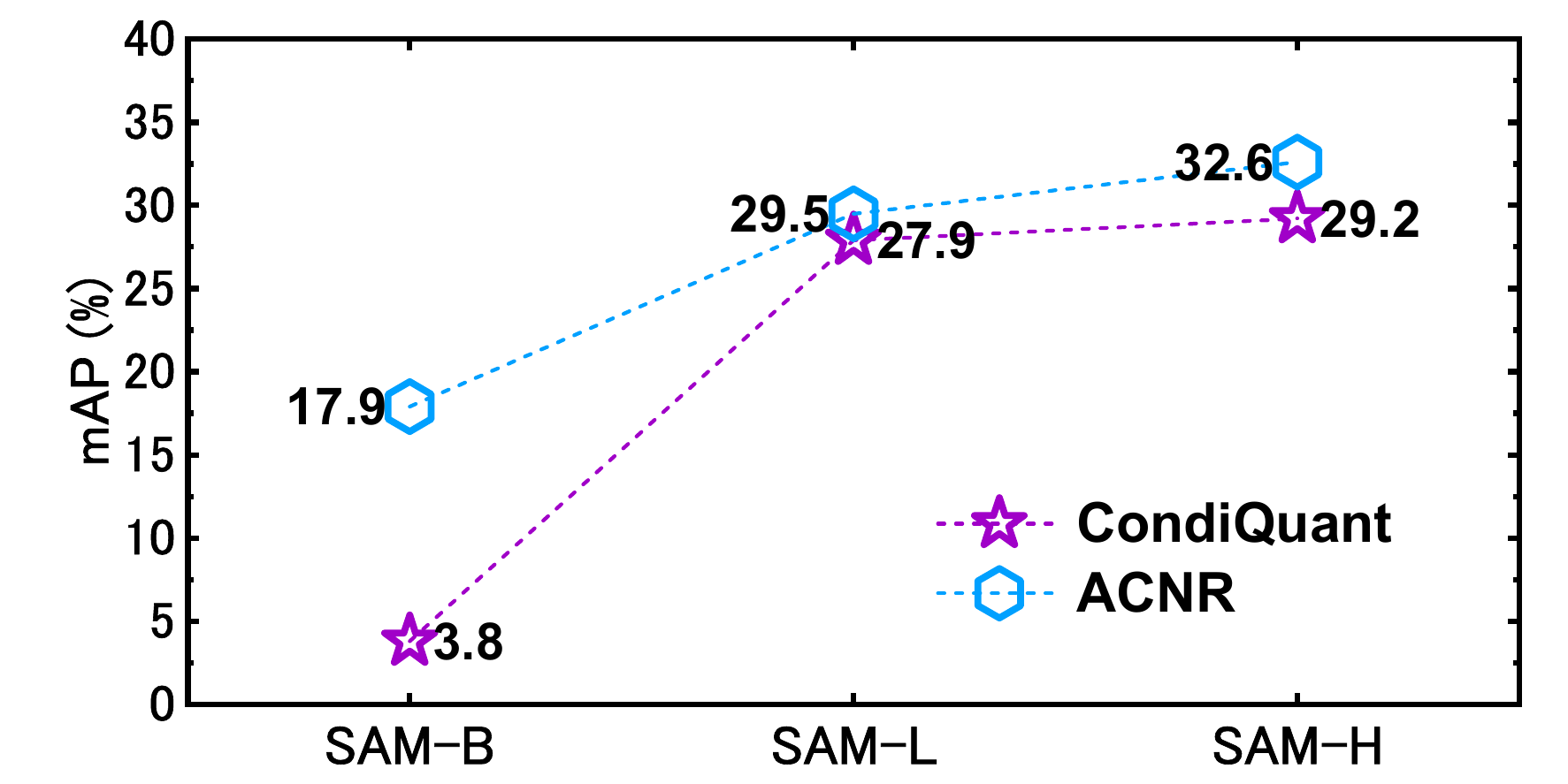}
        \caption{Ablation study of ACNR and CondiQuant. The results are evaluated on 4-bit SAM-B/L/H models with Faster R-CNN.}
        \label{Figure_Framework_Ablation_ACNRandCondiQuant}
\end{figure}

Fig.~\ref{Figure_Framework_Ablation_ACNRandCondiQuant} presents a comparison between the proposed ACNR and CondiQuant~\cite{liu2025condiquant}. As observed, ACNR consistently and significantly outperforms CondiQuant across all SAM variants. Specifically, CondiQuant suffers from severe overfitting, leading to catastrophic performance degradation. For instance, it achieves a mere 0.3\% mAP on SAM-B. In contrast, ACNR substantially recovers the accuracy to 17.9\% mAP. Consistent performance gains are also observed on the larger SAM-L and SAM-H models, underscoring the superior efficacy of our ACNR.

\subsection{Comparison of Visualization Results}

Fig.~\ref{Fig_Visualization} illustrates the visualization results for W4A4 quantization of SAM-H using YOLOX. In comparison with existing methods such as BRECQ~\cite{Brecq}, QDrop~\cite{Qdrop}, and PTQ4SAM~\cite{PTQ4SAM}, AHCQ-SAM consistently generates segmentation masks that are resemble the original floating-point model, preserving fine structural details and sharp object boundaries. These qualitative results demonstrate that AHCQ-SAM effectively mitigates the quantization challenges inherent in SAM, achieving segmentation quality on par with the floating-point baseline. This further confirms its efficacy and robustness for practical low-bit deployment.

\subsection{Hardware Validation}
\label{Subsection_Hardware_Validation}

\begin{figure}[htbp]
    \centering
        \includegraphics[width=\linewidth]{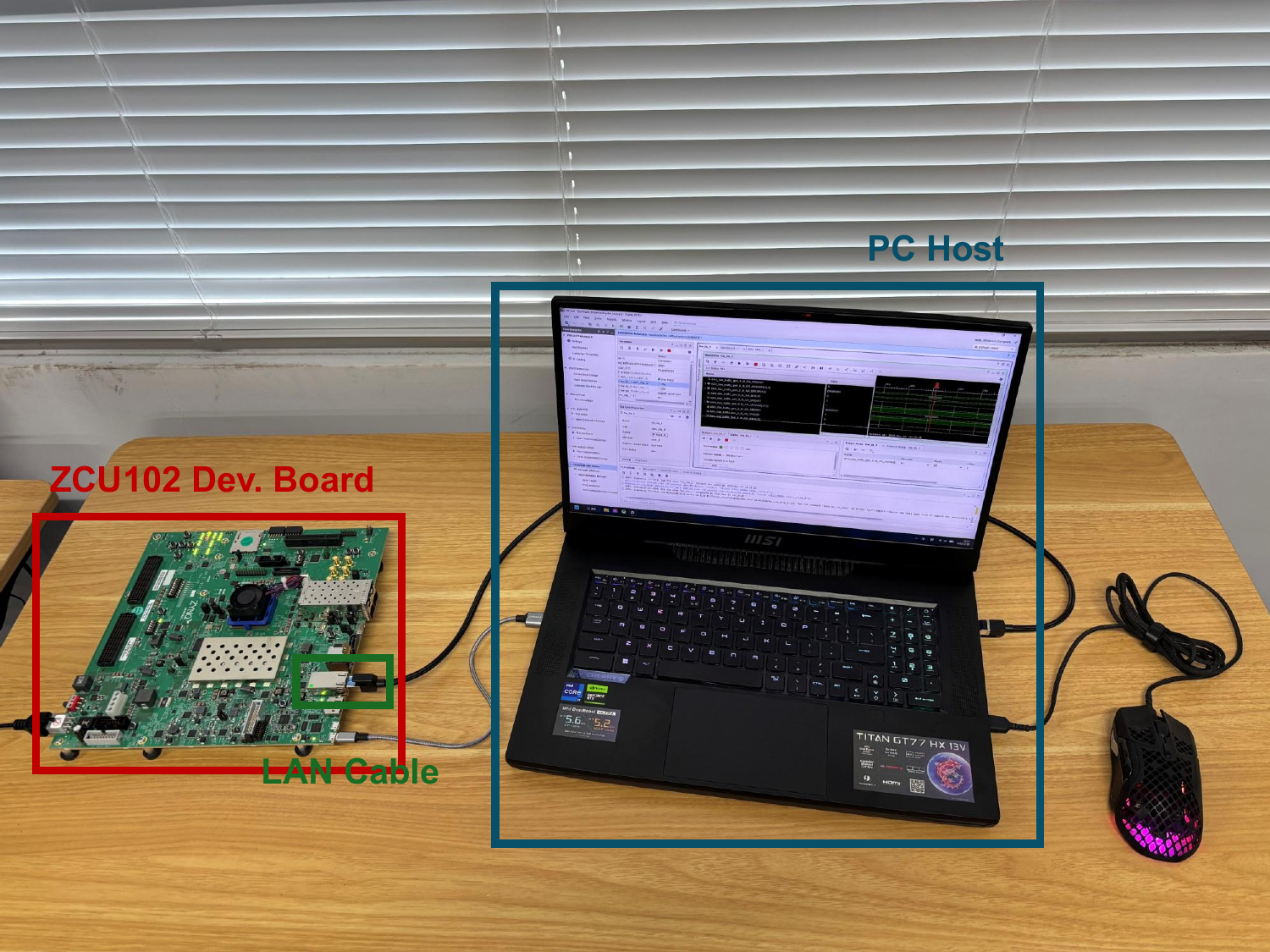}
        \caption{FPGA validation environment.}
        \label{Fig_FPGA_Validation}
\end{figure}

To evaluate the resource efficiency and practical performance of AHCQ-SAM in real-world applications, we developed an FPGA-based accelerator. The accelerator is tailored for SAM-B, where CAG, HLUQ, and LNQ are applied to the corresponding layers, as illustrated in Fig.~\ref{Fig_Framework}. The accelerator is implemented in Verilog, synthesized using Vivado Design Suite, and deployed on an AMD ZCU102 evaluation board operating at 300 MHz. The overall system architecture comprises three components: an FPGA accelerator responsible for large-scale computation, a DDR4 DRAM for data buffering, and a host PC that transfers activations via Ethernet, as depicted in Fig.~\ref{Fig_FPGA_Validation}. For benchmarking purposes, we implemented two baseline accelerators: a standard FP32 accelerator and a default 8-bit integer (INT8) accelerator. Complex arithmetic functions, including Softmax, GELU, and the quantization/dequantization operations of the integer accelerator, are synthesized using Vitis HLS. In contrast, the PEs of the FP32 accelerator are implemented using the Floating-Point Operator IP generator and utilize on-chip DSP resources.

As shown in Tab.~\ref{Table_Resource_Speed}, we report the frame rates and power efficiency of SAM-B when the outputs of Faster R-CNN are used as box prompts. We also summarize the FPGA resource utilization and supported parallelism for each configuration. The FP32 accelerator is limited by the number of available on-chip DSP resources and supports only 16 parallel PE lanes. In contrast, both the default INT8 accelerator and the proposed AHCQ-SAM INT4 accelerator replace DSP-based PEs with LUT-based PEs, enabling a substantial increase in parallelism to 64 and 128 lanes, respectively. Meanwhile, the DSP resources in the integer accelerators are allocated to complex arithmetic operations and the quantization/dequantization processes, improving overall resource utilization and system-level performance. In addition, the BRAM resources remain largely underutilized in the accelerators. We therefore leverage on-chip BRAM as LUTs to simplify the quantization operations in LNQ. The additional LUTs require only 3.2 Mb of BRAM, resulting in negligible overhead. As a result, the 4-bit AHCQ-SAM significantly reduces computational complexity and data movement overhead, achieving $7.12\times$ speedup and $6.62\times$ improvement in power efficiency compared with the floating-point baseline. These results demonstrate the effectiveness of AHCQ-SAM for efficient SAM deployment on edge devices.

\begin{table}[htbp]
\centering
\caption{Analysis of resource utilization and system-level performance of AHCQ-SAM on an FPGA platform.}
\label{Table_Resource_Speed}
\begin{tabular}{c|ccc}
\toprule
\textbf{} & \textbf{FP32} & \textbf{INT8} & \textbf{AHCQ-SAM INT4} \\
\midrule
\textbf{LUT Usage}   & 150,540 & 183,505 & 177,001 \\
\textbf{DSP Usage}   & 1,886 & 453 & 474 \\
\textbf{BRAM Usage} & 253 & 146 & 243 \\
\textbf{Parallelism} & 16 & 64 & 128 \\
\midrule
\textbf{Frame Rate (FPS)}  & 4.75 & 16.34 & \textbf{33.82} \\
\textbf{Power (GOPS/W)} & 7.21 & 25.11 & \textbf{47.73} \\
\bottomrule
\end{tabular}
\end{table}

\section{Conclusion}

This paper presents AHCQ-SAM, a novel PTQ framework designed to effectively address quantization challenges within SAM while ensuring compatibility with hardware acceleration. We first identify four critical challenges that hinder low-bit quantization in SAM: ill-conditioned weight matrices, skewed post-GELU activations, pronounced inter-channel variance, and the exponentially scaled range of attention scores. To overcome these, we introduce four synergistic components, including ACNR, HLUQ, CAG, and LNQ. Specifically, ACNR regularizes weight matrices via a proximal point algorithm to reduce ill-conditioning; HLUQ employs a hybrid log-uniform strategy to capture skewed activations; CAG clusters channels with homogeneous statistics to mitigate inter-channel variance with minimal hardware overhead; and LNQ utilizes logarithmic transformations to adapt to the exponential and heterogeneous attention scores. Experimental results demonstrate that AHCQ-SAM consistently achieves state-of-the-art performance. Furthermore, we establish the PTQ benchmark for SAM2, where AHCQ-SAM also outperforms existing methods, setting a strong baseline for future research. Finally, FPGA-based evaluations confirm its real-world deployability with substantial speedups and power efficiency gains, providing valuable insights for practical applications.

\bibliographystyle{IEEEtran}
\bibliography{main}

\begin{IEEEbiography}[{\includegraphics[width=1in,height=1.25in,clip,keepaspectratio]{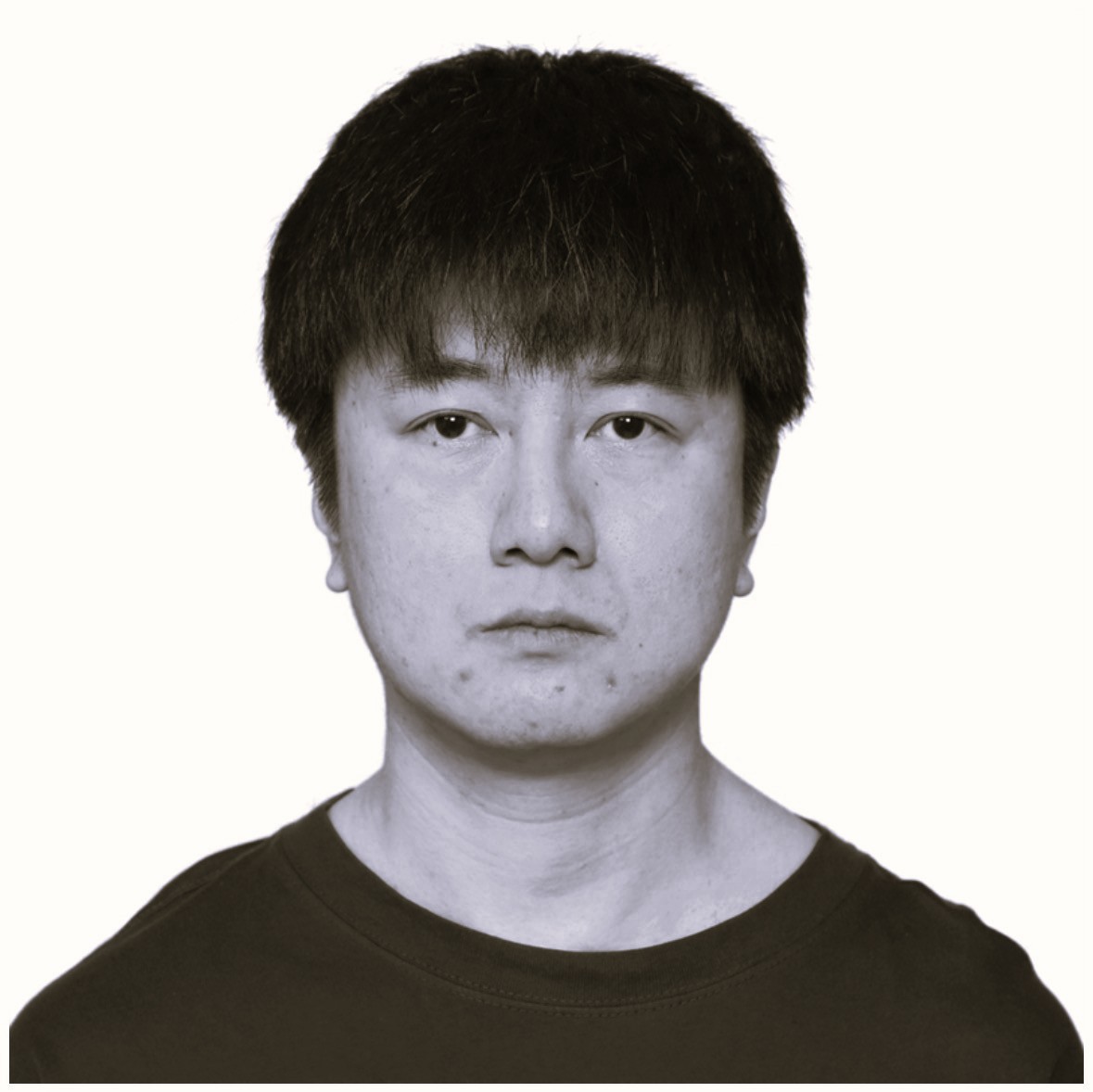}}]{Wenlun Zhang} is currently pursuing the PhD degree with Keio University, Yokohama, Japan. He received the B.E. degree in electrical engineering from Shanghai Dianji University, Shanghai, China, in 2015, and the M.E. degree in electrical and electronic engineering from Tokyo Institute of Technology, Tokyo, Japan, in 2019. From 2013 to 2015, he worked at the Aviation Industry Corporation of China (AVIC), focusing on FPGA design and validation. From 2019 to 2024, he was with Micron Technology, Inc., where he contributed to DRAM circuit design. He has published some papers on top-tier conferences, including CVPR, ICCV, ICCAD, and so on. His research interests include VLSI circuit design, efficient AI, and AI-driven applications. He filed 8 US patents and was the recipient of the PAKDD 2025 Best Paper Award and IEICE VLD Excellent Student Author Award for ASP-DAC 2026.
\end{IEEEbiography}

\begin{IEEEbiography}[{\includegraphics[width=1in,height=1.25in,clip,keepaspectratio]{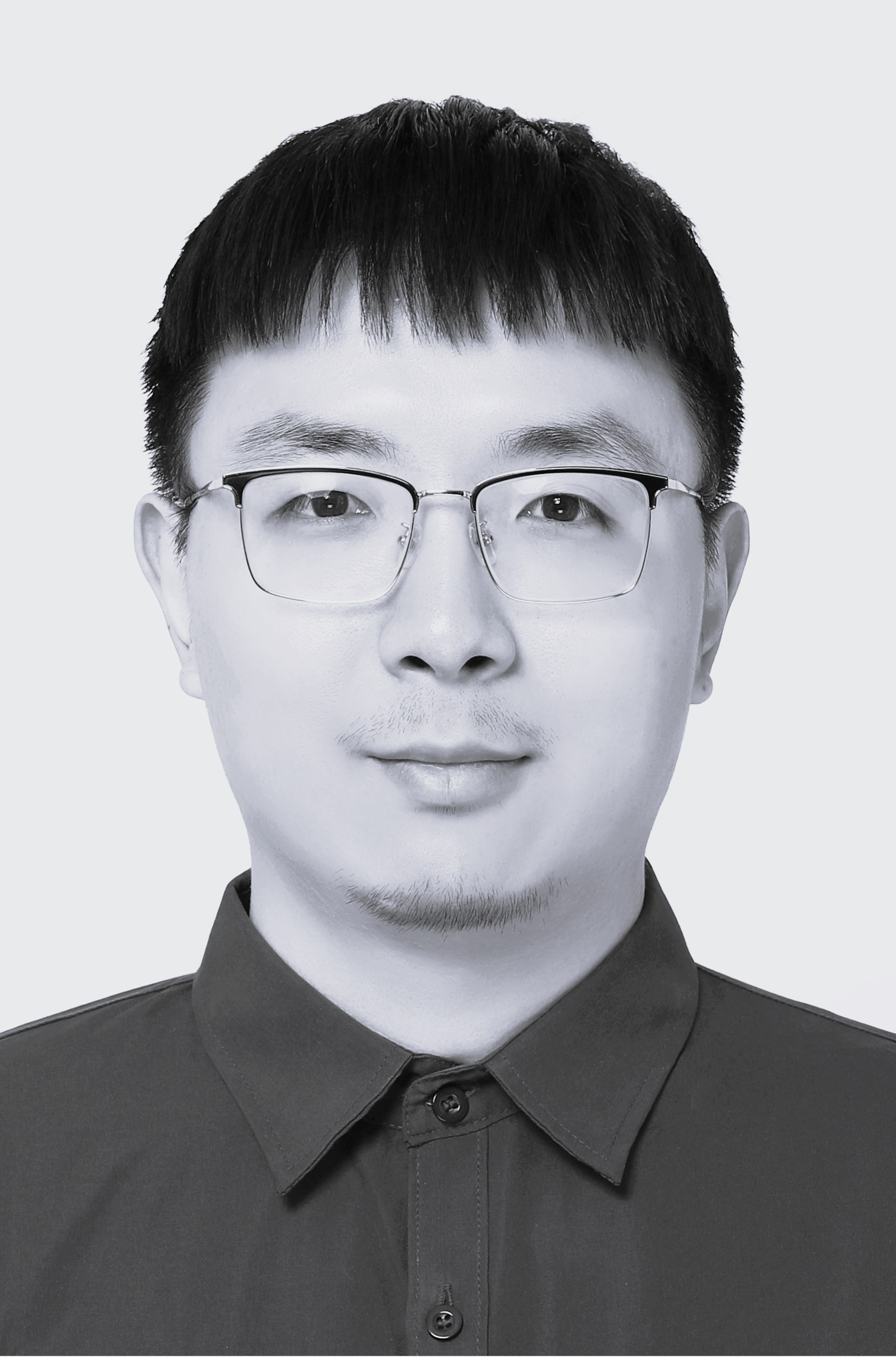}}]{Yunshan Zhong} is an associate professor with Hainan University. He received the B.Sc degree in Software Engineering from Beijing Institute of Technology, Beijing, China, in 2017, the M.S. degree in Software Engineering from Peking University, Beijing, China, in 2020, the Ph.D. degree in the MAC lab, the Institute of Artificial Intelligence, Xiamen University, China, in 2025, under the supervision of Prof. Rongrong Ji. He has published multiple peer-reviewed papers on top-tier conferences/journals, including IEEE TPAMI, ICML, ICLR, CVPR, ICCV, and so on. His current research interest is model compression.
\end{IEEEbiography}

\begin{IEEEbiography}[{\includegraphics[width=1in,height=1.25in,clip,keepaspectratio]{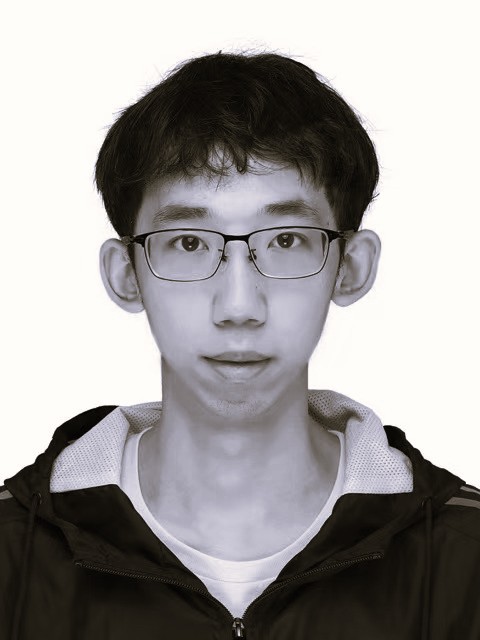}}]{Weiqi Yan}
is currently working towards a PhD degree with Xiamen University, China, under the supervision of Prof. Shengchuna Zhang. His publications on top-tier conferences include CVPR, ICLR, IJCAI, and so on. His research interests include multimodal learning, computer vision, and machine learning.
\end{IEEEbiography}

\begin{IEEEbiography}[{\includegraphics[width=1in,height=1.25in,clip,keepaspectratio]{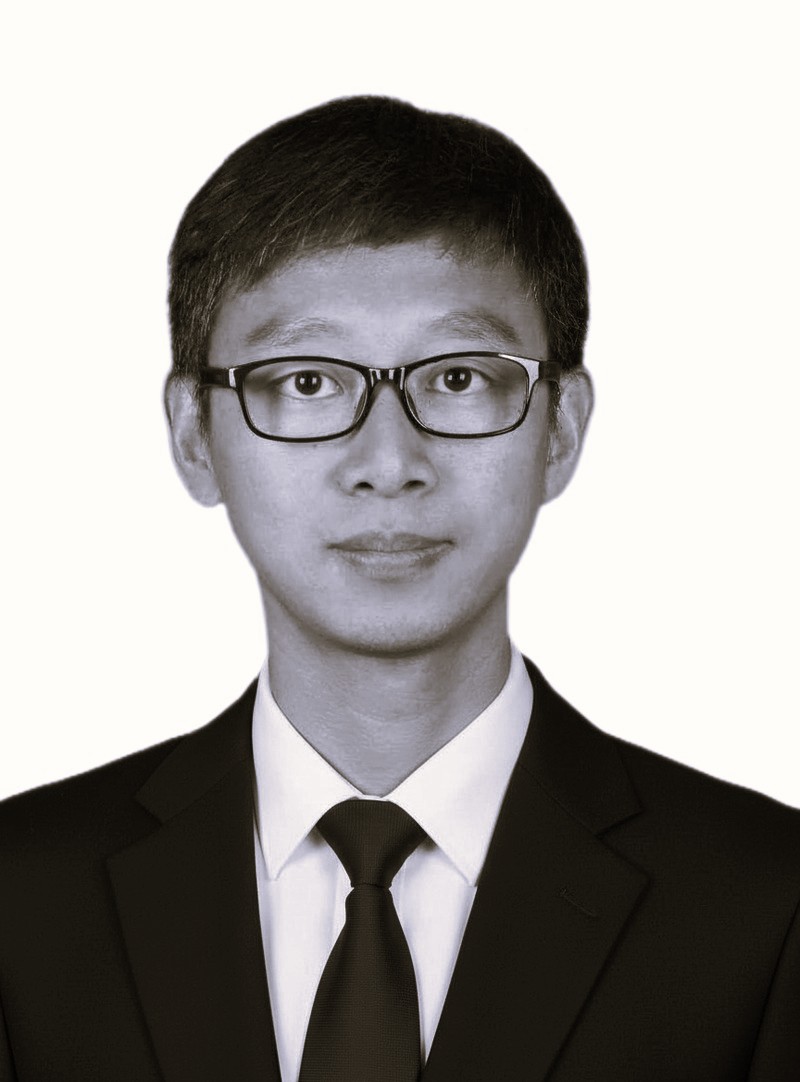}}]{Shengchuan Zhang}
is an associate professor with Xiamen University. 
He received the BEng degree in electronic information engineering from Southwest University, Chongqing, China, in 2011 and the PhD degree in information and telecommunications engineering, School of Electronic Engineering, Xidian University, Xi’an, China, in 2016.
His current research interests include computer vision and pattern recognition. He has published some scientific papers in leading journals, such as IEEE TPAMI, IEEE TIP, IEEE TMM, CVPR, and so on.
\end{IEEEbiography}

\begin{IEEEbiography}[{\includegraphics[width=1in,height=1.25in,clip,keepaspectratio]{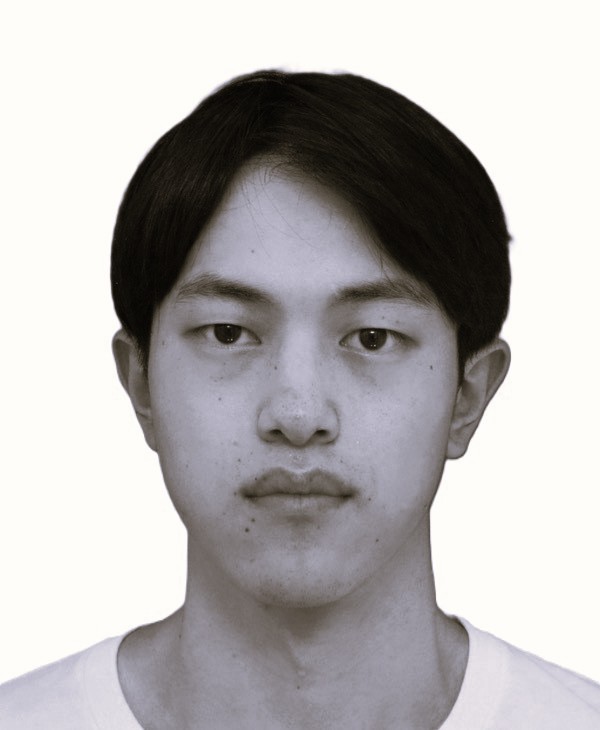}}]{Shimpei Ando}
is currently pursuing the PhD degree at Keio University, Yokohama, Japan. He received the B.S. and M.S. degrees in electrical engineering from Keio University, Yokohama, Japan, in 2023 and 2025. His research interests include Computing In-Memory, Deep Learning, and AI Accelerator.
\end{IEEEbiography}

\begin{IEEEbiography}[{\includegraphics[width=1in,height=1.25in,clip,keepaspectratio]{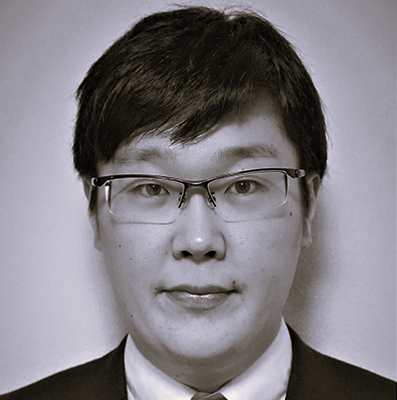}}]{Kentaro Yoshioka}
is currently an Associate Professor at Keio University. He received the B.S., M.S., and Ph.D. degrees from Keio University, Yokohama, Japan. He worked with Toshiba Corporation, Kawasaki, Japan, from 2014 to 2021, developing circuitry for Wi-Fi and LiDAR SoCs. From 2017 to 2018, he was a Visiting Scholar at Stanford University, Stanford, CA, USA, exploring efficient machine learning hardware and algorithms. He has published multiple top-tier conference/journal papers across various fields, including ISSCC, VLSI Symposim, CVPR, ICCV, NDSS, CCS, ICRA, JSSC, and so on. Dr. Yoshioka currently serves as a TPC Member for the IEEE Symposium on VLSI Technology and Circuits. He was the (co-)recipient of the VehicleSec Best Short Paper Award Runner-Up, the CICC Outstanding Student Paper Award, the ASP-DAC Special Feature Award, the A-SSCC Best Design Award, and the First Place Winner of Kaggle 2020 Prostate Cancer Grade Assessment (PANDA) Challenge.
\end{IEEEbiography}

\end{document}